%% file: ms.tex
\pdfoutput=1
\RequirePackage{fix-cm}
\documentclass[smallcondensed]{svjour3}     
\smartqed  
\usepackage{graphicx}
\RequirePackage[colorlinks,citecolor=blue,urlcolor=blue]{hyperref}
\usepackage[T1]{fontenc}
\usepackage[utf8]{inputenc}
\usepackage[english]{babel}
\usepackage[nolist]{acronym}
\usepackage{listings}
\usepackage{url}
\usepackage[FIGTOPCAP]{subfigure}
\usepackage{multirow}
\usepackage{longtable}
\usepackage{amsmath}
\usepackage{mathtools}
\usepackage{breqn}
\usepackage{xcolor}
\DeclarePairedDelimiter{\norm}{\lVert}{\rVert}

\definecolor{lightGrey}{rgb}{0.9, 0.9, 0.9}

\journalname{Journal of Intelligent \& Robotic Systems}
\begin{document}

\title{A Dataset Schema for Cooperative Learning from Demonstration in a Multi-robots System
}%

\titlerunning{Dataset Schema for Cooperative Learning in a MRS}        

\author{Marco A. C. Simões%
    \and%
        Tatiane Nogueira%
    \and%
        Robson Marinho da Silva
}%


\institute{Marco A. C. Simões\href{https://orcid.org/0000-0002-7806-2282}{\includegraphics[scale=1]{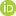}} \at%
              Universidade do Estado da Bahia(UNEB)\\%
              Nucleo de Arquitetura de Computadores e Sistemas Operacionais (ACSO)\\%
              Rua Silveira Martins, 2555, Cabula, Salvador-BA, Brazil\\%
              Tel.: +55-71-999774877\\%
              \email{msimoes@uneb.br} 
           \and%
           Tatiane Nogueira \at%
              Universidade Federal da Bahia(UFBA)\\%
              Programa de Pós-Graduação em Ciência da Computação(PGCOMP)\\%
              Av. Ademar de Barros, s/n, Ondina, Salvador-BA, Brazil\\%
              \email{tatiane.nogueira@ufba.br}%
            \and%
            Robson Marinho da Silva \at%
                Universidade do Estado da Bahia(UNEB)\\%
                Nucleo de Arquitetura de Computadores e Sistemas Operacionais (ACSO)\\%
                Rua Silveira Martins, 2555, Cabula, Salvador-BA, Brazil\\%
                \email{robsonms@uneb.br}%
}

\date{Received: date / Accepted: date}%

\maketitle

\begin{abstract}
\acp{MAS} have been used to solve complex problems which demand  intelligent agents working together to reach the desired goals. 
These Agents should effectively synchronize their individual behaviors so that they can act as a team in a coordinated manner to achieve the common goal of the whole system. 
One of the main issues in \acp{MAS} is the agents' coordination, being common domain experts observing \acp{MAS} execution disapprove agents' decisions. 
Even if the \ac{MAS} was designed using the best methods and tools for agents' coordination, this difference of decisions between experts and \ac{MAS} is confirmed. 
Therefore, this paper proposes a new dataset schema to support learning  the coordinated behavior in \acp{MAS} from demonstration. 
The results of the proposed solution are validated in a \ac{MRS} organizing a collection of new cooperative plans recommendations from the demonstration by domain experts.
\end{abstract}

\input{acronimos.tex}

\acresetall

\section{Introduction}\label{sec:intro}

\acp{MAS} are solutions to complex problems demanding a team of intelligent autonomous agents acting towards their common goals. To achieve these common goals, agents in a \ac{MAS} should be capable of interacting with other agents, not simply by exchanging data, but by engaging as in social activities,  such as those people participate in their daily lives: cooperation, coordination, negotiation, and the like. 
In \acp{MAS},  agents are assumed to be autonomous - capable of making
independent decisions about to do in order to satisfy their design objectives, and thus they need mechanisms that allow them to synchronize and to coordinate their activities at run time \cite{wooldridge_introduction_2002}.
Although one of the main issues in \acp{MAS} is the agents’ coordination
structure, this is not hard-wired at design time, as \acp{MAS} are typically in standard concurrent/distributed systems. 

One well-known strategy for coordination in \ac{MAS} is the design of multi-agent coordinated plans \cite{dambrosio_scalable_2013}\cite{zhang_modeling_2015}\cite{zhang_multirobot_2017}\cite{yu_multiagent_2015}\cite{liemhetcharat_allocating_2017} that include, not only usual agents' actions defined by their effectors, but also communication actions to achieve the necessary synchronization and coordination. To  represent communication actions, some specific languages  were created, e.g.\ \ac{KQML}, \ac{KIF} \cite{wooldridge_introduction_2002}. \ac{KQML} is a message-oriented language that defines a common format for messages to be exchanged between agents in a \ac{MAS}. The development of \ac{KIF} originally intended to provide a common language for expressing properties of a particular domain, i.e., the purpose of \ac{KIF} was to express an ontology for a specific domain. Later, it was also used to represent messages between agents. 

Although the languages and ontologies defined for \ac{MAS} are generally useful, there are some specific domain properties and relations that are relatively hard to be expressed. One of these  domains is \acp{MRS}, often used in stochastic, partially-observed, dynamic environments, and whereby an agent in a \ac{MAS} represents each robot.  One well-known challenge of these complex domains is robotic soccer, as proposed by the RoboCup Federation\footnote{https://www.robocup.org} into the RoboCup Soccer Simulation League\footnote{https://ssim.robocup.org}. 
This league presents a context in which teams composed of eleven agents face each other in a soccer match. 
This challenge is considered a good test-bed for developing coordinated plans  named setplays, in the robotic soccer domain. 

Many works have developed solutions to create setplays in the robotic soccer domain \cite{mota_co-ordination_2010}\cite{fabro_using_2014}\cite{almeida_automatic_2013}\cite{freelan_towards_2014}\cite{bianchi_heuristically_2017}\cite{shi_adaptive_2018}. 
Section \ref{sec:background} presents details about these solutions and tools. 
However, even with all the development in setplays design, situations have commonly been observed whereby domain experts would recommend a different setplay, from the one decided by the team of agents.
This difference is due to the informality present in the specialized knowledge. 
In fact, it is often hard to extract some natural features from domain experts during the knowledge extraction design phase, when an intelligent agent project has been started. For example, in a robotic soccer domain, two different domain experts may recommend different triangulation setplays to overcome an opponent player, but both setplays are equivalent and a \ac{MRS} can consider these setplays as a unique choice to deal with the posed situation.

\ac{LfD} has been used to learn robots basic skills or even very simple setplays in a \ac{MRS} \cite{freelan_towards_2014}. Although, there is no register of using \ac{LfD} to learn complex setplays. Therefore, this work proposes using \ac{LfD} to offer domain experts a chance to watch robotic soccer matches and suggest new setplays for each situation for which they think the \ac{MRS} has made a bad decision. 

Section \ref{sec:rel-work} presents the state-of-the-art for learning coordinated plans in \ac{MAS}.
One of the main issues in \ac{LfD} for setplays is the nature of the dataset generated from the domain experts recommendation. 
Some features in this dataset are not of primitive types as scalars or strings, but some complex types,  such as objects, structures, trees, etc. 
Thus, we also define a strategy presented in Section \ref{sec:problem-definition}, to handle this kind of complex data. 

The proposed solution has a two-level dataset, detailed in Section \ref{sec:dataset}. To assess the feasibility of using this dataset to support setplays learning, the \ac{FCM} algorithm is used to organize setplays recommendations into clusters. 
The choice of the \ac{FCM} algorithm is due to the imprecision inherent in the friendly interface proposed for use by experts to generate the recommendations of setplays. 
The suggestions from experts are organized in clusters to solve the semantic equivalence issue presented in Section \ref{sec:background}. 
Section \ref{sec:assessment} describes the assessment process and its results. 
Section \ref{sec:conclusion} has conclusion and future work descriptions.

\section{Background}\label{sec:background}

In a \ac{MRS}, an agent always faces a situation defined by the properties extracted from the environment in the current or previous states. This set of features states how the agent sees the world, which is called \textit{world state} or \textit{mind state}. Hence, a sequence of situations summarizes the whole agent life. These situations can be strategic or active. Whenever the agent is positioned to favor the team behavior, there is a strategic situation going on. Otherwise, when an agent performs active behaviors (e.g.\ passing a ball, dribbling a ball, intercepting a ball, etc) that generates relevant changes in the environment state,  an active situation is going on \cite{reis_situation_2001}.

We here focus on active situations. When an agent detects an active situation, it can either decide to perform a single active behavior or it can choose to fire an instance of a predefined cooperative plan (setplay). Each setplay has an initial condition $C_I$ that identifies the situations in which this setplay may be activated.

A setplay can be defined as a multi-step flexible plan involving a variable number of robots \cite{mota_multi-robot_2011}. Originally, setplays were defined as a combination of \cite{stone_task_1999}:

\begin{itemize}
    \item an activation condition that defines a set of world states in which the setplay will be activated;
    \item a set of $m$ setplay roles $R_{sp} = \{spr_1, \ldots, spr_m\}$ that defines the behaviors that to be executed by the robots participating in the setplay. Each setplay role $spr_i$ ($i=1,\ldots,m$) includes:
    \begin{itemize}
        \item a behavior to be executed; and 
        \item a end condition indicating a set of states of the world in which the agent must abandon its role in the setplay and resume its normal behavior.
    \end{itemize}
\end{itemize}

This definition limits the setplay to a single flow of steps. At first, setplays were used in specific situations, such as dead-ball plays (e.g.,\ corner kicks, kick-ins, etc.) in robotic soccer. Then, the concept was extended to support multiple streams of steps and applications to various scenarios \cite{mota_multi-robot_2011}.

A setplay must have a unique name or identification and a list of references to the agents taking part in a specific setplay activation. These references can be related to specific agents or roles in the team strategy definition \cite{reis_situation_2001}. A set of abort conditions defines the world states which lead to setplays abortion. The setplay roles were renamed as \textit{steps} and their concept was extended \cite{mota_multi-robot_2011}. Steps are fundamental blocks used to build setplays. Setplays are built from a list $S=\{S_1, S_2, \ldots, S_n\}$ containing $n$ steps. A step represents a state during a setplay activation. Agents participating in a setplay can be part of all its steps or just a few. Each step has an unique numerical \textit{id}. A step is defined as:
\begin{itemize}
    \item a list of behaviors $B_s=\{b_1, \ldots, b_{\kappa}\}$, whereby $s$ is the step \textit{id} and $\kappa$ is the number of agents participating in step $s$;
    \item a set of transition conditions $\tau_s=\{C_1, \ldots, C_k\}$, whereby $k$ is the number of transitions in step $s$
\end{itemize}

A specific step $0$ is the initial step by default. The transition conditions in $\tau_s$ are based on the properties from the agents' world state. Appropriate temporal transition conditions may also be part of $\tau_s$. \textit{Wait time} is a special condition that defines the minimum time a set of agents should wait to perform a transition to another step. \textit{Abort time} is the maximum time a set of agents can stay at the same step. If the \textit{abort time} is reached, the setplay execution is aborted. A subset of $\tau_s$ defines special terminal conditions $C_{suc}, C_{f1}, C_{f2}, \ldots, C_{fp} \in \tau_s$, where $suc,f1,f2,\ldots,fp<k$. $C_{suc}$ is a success condition that indicates the setplay has successfully finished. The $C_fj, 1<j<p,$ are failure conditions, that indicate the setplay has unsuccessfully finished. Not all steps contain success or failure conditions in their $\tau_s$ set.

\begin{figure}[htb]
    \centering
    \includegraphics[width=.6\textwidth]{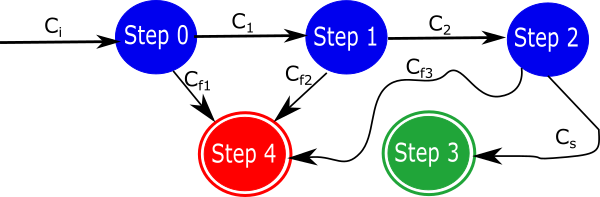}
    \caption{Setplay modeled as a \ac{DFA}. Steps  $0, 1$ e $2$ are the usual setplays steps, including the initial step $0$. Step $3$ is the final step, which indicates successfully finishing the setplay. Step $4$ indicates that the setplay aborts with failure. Condition $C_i$ is the initial condition that indicates the situation in which the \textit{setplay} is activated. $C_1$ e $C_2$ are transition conditions and $C_s, C_{f1}, C_{f2}$ e $C_{f3}$ are terminal conditions.}
    \label{fig:setplay-dfa}
\end{figure}

A setplay can be modeled as a \ac{DFA} in which steps are represented by states and the transitions in $\tau_s, \forall s \in S,$ are modeled as the state transitions in a \ac{DFA}. Figure \ref{fig:setplay-dfa} illustrates an example of a setplay modeled as a \ac{DFA}.

One agent participating in a step never changes its behavior while keeping the same current step. Thus, the set of transition conditions $\tau_s$ of a step $s$ should include all the conditions that fire a transition if at least one of the participating agents changes its current behavior.

Agents' coordination in a setplay should be achieved using a defined communication and synchronization policy \cite{mota_multi-robot_2011}. The main goal is to assure that all the agents synchronize to the same setplay step. Considering that information in the world model of each agent is not equal, it would be possible for some agents to define one current step and other agents believe they are at different current step. Communication enables the synchronization of steps, which is an essential requirement for the setplay to work. The communication policy defines one agent at each step to be the step leader. The same agent can be the leader of more than one step. In the robotic soccer domain, for example, the agent that takes the ball possession can be the agent leader. In defensive setplays, the closest agent to the ball can be the step leader. Once a clear and unambiguous criterion is chosen to define the step leader, the policy will work to synchronize the steps during setplay execution. 

The leader agent broadcasts messages to all the agents with information about the current setplay, the current step, the transition conditions, and abort conditions. Therefore, only the leader agent decides about steps transitions, and all the agents are synchronized with this decision.

Although setplays are an essential solution for developing \acp{MRS} capable of solving complex problems by coordinating simple behaviors, hand-coding a setplay can be an arduous task. The next subsection describes some tools developed to facilitate the task of developing a setplay using high-level languages more abstract than programming languages. 

\subsection{Tools for designing setplays}\label{sec:tools}

\ac{FSF} is a free software to support setplays building using collaborative behaviors in a team of robotic soccer players \cite{hutchison_collaborative_2015}. \ac{FSF} uses a grammar based on S-Expressions to define an abstract language to express setplays in the problem domain abstraction level. This language can be used to interpret and to execute setplays in any soccer competition, even in human soccer matches. Setplays can be used in dead ball situations or regular play-on mode. 

Listing \ref{cod:setplaybegin} shows a fragment of a setplay expressed in the S-expression language defined by \ac{FSF}. A name and a unique id identify the setplay~(line \ref{line:nameid} in listing~\ref{cod:setplaybegin}). Other important general properties are the list of players taking part in the setplay~(lines~\ref{line:playersbeg}--\ref{line:playersend} in listing~\ref{cod:setplaybegin}) and the general abort conditions~(line~\ref{line:abort} in listing~\ref{cod:setplaybegin}). In this example, the abort condition is the ball owner~(\textit{bowner}), one of the players in the opponent team~(\textit{:team opp}) and the current play-mode is not \textit{play\_on} or kick off to our team~(\textit{ko\_our}). When this condition arises in any step, the setplay aborts. 

The code fragment in Listing \ref{cod:setplaybegin} was generated using \ac{SPlanner} \cite{hutchison_collaborative_2015}. \ac{SPlanner} is a graphical tool that provides the domain expert with an easy to use interface to create setplays. The \ac{SPlanner} user does not need to understand the S-expressions syntax. It interacts only with a \ac{DFA} representing the setplays steps and a soccer field where it puts the players and defines their behavior for each step. When the setplay finishes, it can be exported to an S-expressions format.

\begin{lstlisting}[caption=Fragment of a text file containing the beginning of a setplay represented in the S-expressions language,label=cod:setplaybegin,breaklines=true,breakatwhitespace=true,escapechar=|]
(setplay :name newSetPlay :id 1 :invertible true |\label{line:nameid}|
    :version splanner_1.5
	:players |\label{line:playersbeg}|
	(list
		(playerRole :roleName Player7)
			(playerRole :roleName Player5)
			(playerRole :roleName Player6)
			(playerRole :roleName Player8)
		) |\label{line:playersend}|
	 :abortCond (or (bowner :players  (list (player :team opp :number 1) (player :team opp :number 2) (player :team opp :number 3) (player :team opp :number 4) (player :team opp :number 5) (player :team opp :number 6) (player :team opp :number 7) (player :team opp :number 8) (player :team opp :number 9) (player :team opp :number 10) (player :team opp :number 11) )) (and (not (playm play_on)) (not (playm ko_our)))) |\label{line:abort}|
\end{lstlisting}

The sequence of steps that compose a setplay is represented using the same syntax as verified in the example of Listing~\ref{cod:setplaysteps}. A step has a unique numeric id and optional abort conditions, e.g.\ \textit{abortTime}~(line~\ref{line:abortTime} in listing~\ref{cod:setplaysteps}). Then, a list of the participants in this step is presented with their positions in the field~(lines~\ref{line:partbeg}--\ref{line:partend} in listing~\ref{cod:setplaysteps}), for example,  at \textit{( playerRole : roleName Player7 ) ( pt : x 0 : y 0)} means \textit{Player7} is in position $(0,0)$. This list must be a subset of the list of players participating in the setplay presented in Listing \ref{cod:setplaybegin}. The players can be identified by an absolute numeric id or by a role in the team strategy. The \textit{:condition} expression states the condition of moving from a previous step to the current one~(line~\ref{line:cond} in listing~\ref{cod:setplaysteps}). If the current step is the initial step ($id=0$), this is the start condition used to fire the setplay. The \textit{:leadPlayer} identifies the leader agent in the current step~(line~\ref{line:lead} in listing~\ref{cod:setplaysteps}).  

\begin{lstlisting}[caption=Fragment of a text file showing a step and its transition in a setplay represented in the S-expression language,label=cod:setplaysteps,breaklines=true,breakatwhitespace=true,escapechar=|]
:steps
 (seq
 (step :id 0 :waitTime 0 :abortTime 26 |\label{line:abortTime}|
    :participants |\label{line:partbeg}|
	  (list
		(at (playerRole :roleName Player7) (pt :x 0 :y 0))
		(at (playerRole :roleName Player5) (pt :x -6 :y -1))
		(at (playerRole :roleName Player6) (pt :x -1.5 :y 3))
		(at (playerRole :roleName Player8) (pt :x -1.5 :y -4))
      )|\label{line:partend}|
	:condition (playm ko_our) |\label{line:cond}|
	:leadPlayer (playerRole :roleName Player7)|\label{line:lead}|
	:transitions |\label{line:transbeg}|
	 (list 
	  (nextStep :id 1
		:directives
		  (list
			(do 	:players (list (playerRole :roleName Player6) )
				:actions (list (pos :region (pt :x 4 :y 3)) )
			)
			(do 	:players (list (playerRole :roleName Player7) )
				:actions (list (bto :players          (list (playerRole :roleName       Player5) ) :type normal) )
			)
			(do 	:players (list (playerRole :roleName Player8) )
				:actions (list (mov :region (pt :x 3.5 :y -4)) )
			)
			(do 	:players (list (playerRole :roleName Player5) )
				:actions (list (intercept) )
			)

		  )
	 )
   )|\label{line:transend}|
 )
\end{lstlisting}

The \textit{:transitions} contain a list of behaviors executed by each participating agent to change the current world state to a condition that fires a transition to the next step~(lines~\ref{line:transbeg}--\ref{line:transend} in listing~\ref{cod:setplaysteps}). The \ac{FSF} must map all the possible behaviors developed in the agents code, in which \ac{FSF} is used. In this example, we can see some behaviors as \textit{mov} and \textit{pos} which are different kind of movements towards a specific region in the field, \textit{bto}, which means kicking the ball in the direction of another player and \textit{intercept} which means trying to intercept the ball. 

The setplay S-expression language defined by \ac{FSF} provides an abstraction to define and to discuss setplays considering only the domain knowledge concepts. Detail from agents implementation is not considered. If a \ac{MRS} needs to use \ac{FSF}, the agents in the \ac{MRS} must extend the abstract classes in the \ac{FSF} and implement their virtual functions. This procedure is the way a \ac{MRS} can interpret and execute a setplay described by the \ac{FSF} S-expression abstract language.

\begin{figure}[htb]
    \centering
    \includegraphics[width=.6\textwidth]{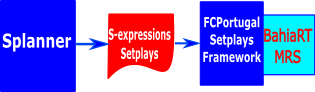}
    \caption{SPlanner generates a S-expression file containing a setplay to be interpreted by \ac{FSF} and executed by \ac{BahiaRT} \ac{MRS}.}
    \label{fig:setplay-tools}
\end{figure}

An initiative to foster scientific development in \ac{AI} and Robotics according to the RoboCup Federation directives is \ac{BahiaRT}\footnote{See http://www.acso.uneb.br/bahiart}, which already extended the \ac{FSF}. Thus, \ac{BahiaRT}'s agents are able to interpret and execute setplays defined in  S-expression language \cite{ramos_planejador_2017}.  The output of \ac{SPlanner} is used as input for the \ac{BahiaRT}'s \ac{MRS} agents which extends the \ac{FSF}. The primary purpose of the initiative \ac{BahiaRT} is to validate research results in practice using scientific competitions such as RoboCup. The interaction of these tools is illustrated in Figure \ref{fig:setplay-tools}.

\begin{figure}[htb]
    \centering
    \includegraphics[width=.8\textwidth]{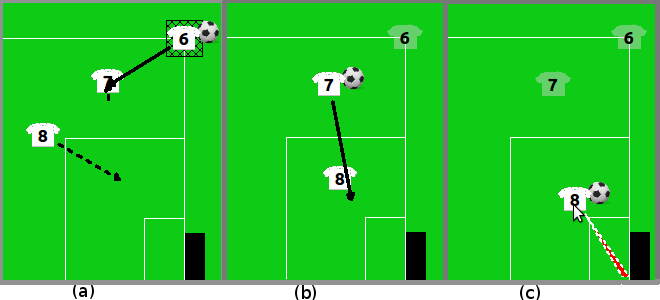}
    \caption{An example of a setplay designed using \ac{SPlanner}. (a) In the first step, agent number 6 performs a pass to agent number 7 while agent number 8 goes into the goal area; (b) in the second step, agent number 7 will performs a pass to agent number 8; (c) in the third step, agent number 8 will try a shot on goal \cite{hutchison_collaborative_2015}}
    \label{fig:splanner-example}
\end{figure}

A complete example of a setplay designed using SPlanner is exhibited in Figure \ref{fig:splanner-example} that shows an offensive corner kick situation. The setplay finishes successfully if the team scores a goal. The user can create the setplay just dragging-and-dropping the players to desired positions and right-clicking on the players to choose one of the behaviors available.  

Using this set of tools, a domain expert can devise the setplay design concerned only with the domain issues. He does not need to worry about programming languages, data structures, or s-expression details. Even supported by these tools, the design of setplays to deal with important situations in a \ac{MRS} is hard to be accomplished. In robotic soccer competitions, setplays designers are usually unsatisfied with their robots cooperative behavior because they think they could perform a different setplay in that situation. 

An option to fill this gap between what setplays designers think and what the agents in the \ac{MRS} effectively do is using machine learning strategies. In a previous work \cite{simoes_towards_2018}, general architecture for learning new setplays from domain experts demonstration was presented. One of the main issues in this proposal is to define an adequate dataset to feed the machine learning engine. This work is concerned to provide such a dataset which can support the machine learning solution. Section \ref{sec:dataset} presents our proposal to solve this problem. Before describing the solution, next subsection defines the concept of setplays semantic equivalence. 

\subsection{Semantic equivalence}\label{sec:semantic-equivalence}

\begin{figure}[htb]%
    \centering
    \subfigure[FIGBOTCAP][ ]{\label{fig:equiv-semantica-a}\includegraphics[width=0.47\textwidth]{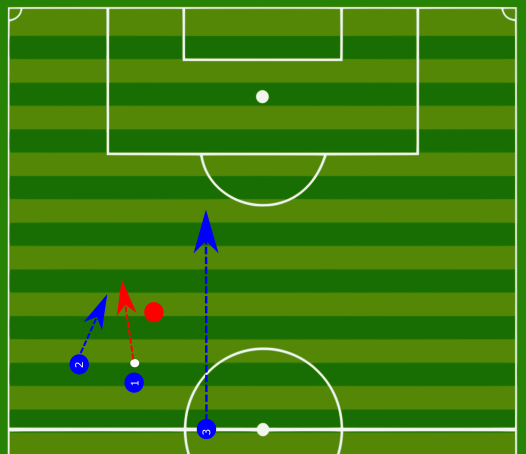}}
    \qquad
    \subfigure[FIGBOTCAP][ ]{\label{fig:equiv-semantica-b}\includegraphics[width=0.47\textwidth]{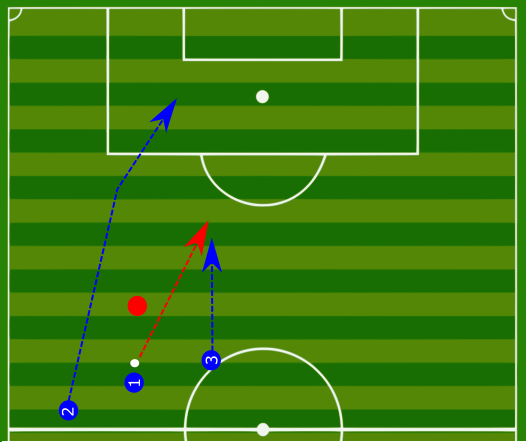}}
    \caption{Semantic equivalence of setplays in the robotic soccer domain.}
    \label{fig:equiv-semantica}
\end{figure}

Domain experts may see two different setplays implemented by a \ac{MRS} and think they represent the same setplay for the domain environment where the \ac{MRS} runs. Let us consider an example in the robotic soccer domain. Figure \ref{fig:equiv-semantica} shows two different setplays.

In the setplay illustrated in Figure \ref{fig:equiv-semantica}\subref{fig:equiv-semantica-a}, the player with ball possession (number $1$) chooses to pass the ball to player number $2$. At the same step, player number $3$  advances. In the remainder of the setplay, player number $2$ receives the pass and performs another pass to player number $3$. The setplay thus finishes successfully when player number $3$ receives the ball. The setplay goal is to overcome the opponent represented by the red circle. If we draw lines joining the two target points of the passes and the initial position of the ball, we  form a triangle. This triangle leads soccer experts to call this kind of play triangulation.

In Figure \ref{fig:equiv-semantica}\subref{fig:equiv-semantica-b}, the setplay described is different from the first setplay for the agents in the \ac{MRS} perspective. For a soccer domain expert, both are triangulations and represent the same soccer play. The differences are the teammate chosen to receive the first pass and the consequent movements performed by the other players to complete the setplay. For a soccer domain expert, both setplays have the same goal (i.e.,\ to overcome the opponent player) and are triangulations. 

If we need to extract this kind of abstract knowledge from a domain expert and translate it to a setplay that a set of agents in a \ac{MRS} can execute, we need to represent this kind of \textit{semantic equivalence} in the dataset used to learn the new setplays.

\begin{definition}[Semantic equivalence]
Two setplays $\sigma_p$ and $\sigma_q$ are considered semantic equivalents if they represent the same play in the domain abstract knowledge level. 
\end{definition}

A challenge derived from this definition is how we can model this concept for a machine learning engine. Let us consider the S-expression fragments presented in Listings \ref{cod:setplaybegin} and \ref{cod:setplaysteps}. If the role names of the players participating in two setplays are different, but the number of steps and the behaviors assigned to each player at all steps are the same, and the abort and finish conditions are equal, we may probably consider the two setplays semantic equivalents. The execution by the \ac{MRS} can be slightly different, but for a domain expert it represents the same play. 

In another example, if in two different setplays the position of players in each step is not the same, but all the players present the same behavior in all steps and the abort and finish conditions and number of steps are equal, the setplays are probably semantic equivalent. This equivalence is a typical case where the same play can be executed in different field regions, but representing the same coordinated plan.

These examples show that there are many situations whereby slightly different setplays are semantic equivalent. The criteria to define if two or more setplays are semantic equivalent are not precise. A clustering engine can organize all the instances of a dataset into groups of semantic equivalent setplays. Due to the imprecision inherent to the semantic equivalence concept, one setplay instance can be a member of more than one group at the same time. In this case, this instance presents different membership degrees to each group.  Section \ref{sec:dataset} presents a fuzzy solution to define setplay clusters  representing the semantic equivalence. Before this, in the following Section, the RoboCup 3D Soccer Simulation used as the source of data for this work is described.

\subsection{The RoboCup 3D Soccer Simulation League}\label{sec:3d-architecture}

The robotic soccer challenge proposed by the RoboCup Federation is a great standard problem to model a \ac{MRS} in which the use of setplays can be validated.
The RoboCup Federation splits soccer challenge into five leagues\footnote{https://www.robocup.org/domains/1}. 
The leagues differ in the kind of robots (physical or simulated, humanoid or wheeled, etc.),  size of the field, the number of players on the field, adaptations to real soccer rules. 
As setplays regard high-level cooperative \ac{AI}, the simulation league is one of the most promising leagues to test setplays.

Soccer Simulation League comprises two sub-leagues: 2D simulation league and 3D simulation league. In the 2D league,  the simulation uses a kind of long-term futuristic robot. This robot has nearly perfect skills, such as running and kicking. The agents are not supposed to control joints or individual body parts. They only need to send commands such as \textit{dash}, \textit{kick} or \textit{tackle} to the simulator, and the simulation server performs an almost perfect behavior. It is an excellent test-bed for setplays. Setplays that are well succeeded in 2D simulation league are not feasible to be executed by current real robots in most cases. No robots are very likely to perform skills similar to 2D robots in the coming years.  

\begin{figure}[htb]
    \centering
    \includegraphics[width=.8\textwidth]{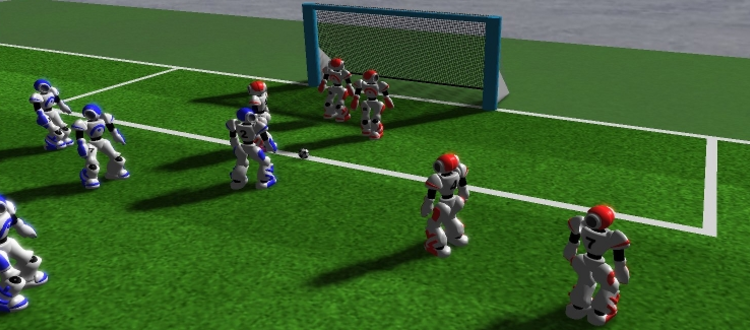}
    \caption{Simulated NAO-based robots playing soccer in the field provided by the Simulation server.}
    \label{fig:nao-field}
\end{figure}

The 3D Soccer Simulation League was created to reduce the gap between simulated robots and real robots. The simulation server models a modified version of a real robot. The NAO robot from Softbank Robotics\footnote{https://www.softbankrobotics.com/emea/en/nao} is the base for the robot model in the simulator. Figure \ref{fig:nao-field} shows the simulated robots playing soccer in the 20x30 meters field provided by the simulation server. Each team uses eleven players, and rules are similar to official soccer rules.

The RoboCup 3D Soccer Simulation league maintains the simulation server as a free software project. Simspark and Rcssserver3d are the two main modules of the simulation server\footnote{https://gitlab.com/robocup-sim}. The Simspark is a generic physics simulator that can simulate any robot model to be used in any application. The Rcssserver 3D is the soccer server that provides specific domain rules, such as soccer rules. The recommended system architecture should use one dedicated computer to run the simulation server Simspark.  

\begin{figure}[htb]
    \centering
    \includegraphics[width=.8\textwidth]{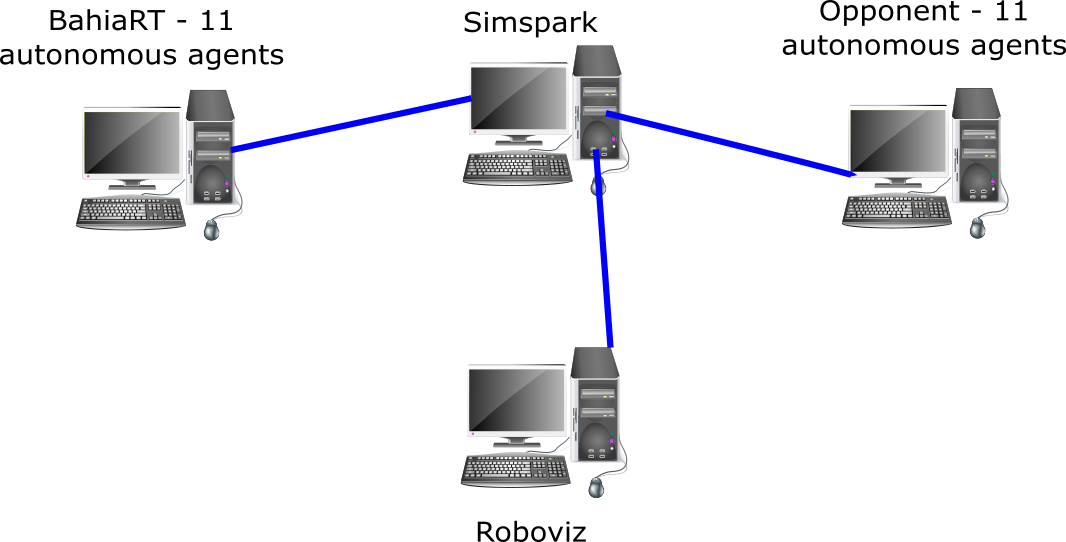}
    \caption{3D Soccer Simulation league system architecture.}
    \label{fig:robocup-3d-architecture}
\end{figure}

Each team is an autonomous \ac{MRS}; each simulated robot is thus controlled by an autonomous agent. 
The eleven agents join a \ac{MAS} running in a dedicated client computer. RoboViz\footnote{https://github.com/magmaOffenburg/RoboViz} is a visualization tool that allows watching the soccer matches. 
RoboViz supports both real-time and recorded matches exhibition. 
In the standard mode, RoboViz connects to the server using local computers network to exhibit real-time matches. When RoboViz launches in \textit{logMode} it can exhibit a match recorded in a log file previously generated by Rcssserver3d and Simspark. 
Figure \ref{fig:robocup-3d-architecture} shows the system architecture for a 3D Soccer Simulation league setup in which \ac{BahiaRT} faces opponent teams in matches simulated by Simspark and exhibited by RoboViz.

\subsection{Problem Definition}\label{sec:problem-definition}

The previous subsections presented existing tools and concepts to design cooperative behaviors in \acp{MRS}. First, complete architecture to learn new cooperative plans (e.g.\ setplays in robotic soccer) from domain experts demonstration was presented \cite{simoes_towards_2018}. Learning setplays means generalizing two kinds of knowledge: 
\begin{itemize}
    \item situations in which setplays are fired; and
    \item setplays components, such as a list of players and their positions, ball position, list of steps and their transitions, abort, and success conditions.
\end{itemize}

Agents world model properties define the situations whereby each setplay may be fired. The agents obtain the values of these properties from their sensors. To define the set of features required to model the situations capable of firing setplays, the current decision-making policy in the \ac{MRS} of team \ac{BahiaRT} was considered. All the features used in the current policy were mapped and organized into a dataset. Data from matches between team \ac{BahiaRT} and teams that participated in the RoboCup 2017 were used to feed the dataset. The learning process aims to define which behavior or setplay each agent should execute depending on the combined values of the features in the dataset. This goal characterizes a classification problem. A decision tree with the CART algorithm \cite{breiman_classification_2017} was used to classify the situations \cite{simoes_towards_2018}.   

The accuracy of the learned classifier was greater than $98\%$ when compared with current \ac{BahiaRT} decision making policy. These results validated the set of $24$ features extracted from the agents world model to describe the situations whereby setplays are fired \cite{simoes_towards_2018}. 

\begin{figure}[htb]
    \centering
    \includegraphics[width=.8\textwidth]{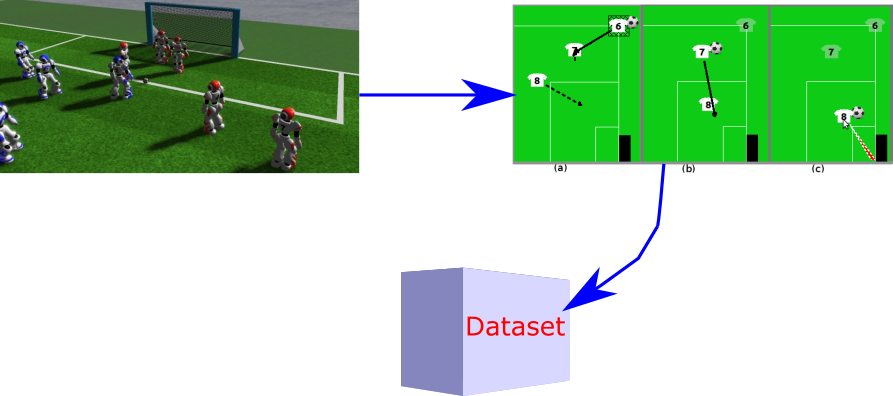}
    \caption{Learning setplays from demonstration. The domain expert uses a modified version of RoboViz to define the starting situation to the recommended setplay. The remaining steps are designed in the SPlanner. The resulting setplay is converted into a recommendation to the dataset for learning new setplays.}
    \label{fig:learning-tools}
\end{figure}

\ac{LfD} directives inspire the proposed solution. The domain expert uses a modified version of RoboViz using a new \textit{demonstration mode}. In this mode, the user can stop a match reproduced from a log file and starts a setplay recommendation beginning in the current match scene. The user can select the teammates and opponent players that take part in the recommended setplay. RoboViz writes a setplay file using the S-expression syntax and launches SPlanner loading this file. The domain expert can use SPlanner starting with a situation derived from the real scene watched in RoboViz. The remaining steps of the setplay can be designed as usual in the SPlanner. The setplay designed must be transformed into a recommendation into a dataset for setplays learning. Figure \ref{fig:learning-tools} illustrates the recommendation process. 

The main problem investigated herein is the dataset generation. As described in Section \ref{sec:background}, the setplay is composed of some non-primitive data types. For example, all the setplays contains Boolean conditions (e.g.\ transitions, abort and finish conditions) and lists of objects (e.g.,\ lists of players and lists of steps). These data types are not scalar types or even strings of characters. As far as we know, all machine learning strategies require pre-processing of non-scalar data to feed their learning engines with an adequate dataset. 

In setplays definition, the data is also hierarchical. For example, a setplay contains a list of steps,  each step contains a list of participating players, and each player contains a pair of coordinates. Semantic equivalence also needs to be considered when solving the dataset generation problem. The \ac{MRS} needs to generalize groups of setplays and not necessarily individual setplays. Since, the knowledge is in the domain experts' level, including its possible informality and imprecision, the bounds between setplays groups are not strict.  Section \ref{sec:dataset} presents a solution to this problem detailing the features in the dataset for a possible clustering approach that generates clusters of setplays to the \ac{MRS}. The next Section discusses some related works.

\section{Related Works}\label{sec:rel-work}

A number of researchers have investigated planning in \ac{MAS}. A scalable solution to large teams ($~1024$ agents) considers the team geometric pattern instead of individual agents positions for multiagent learning \cite{dambrosio_scalable_2013}. A neural network model named HyperNEAT was used to generalize the agents roles in the \ac{MAS} from the agents positioning in the system. The authors argue that their contributions are conceptual. The model was not validated in real applications or well-known challenges in the scientific community. 

Another work \cite{zhang_modeling_2015} describes the use of Bayesian networks to learn the behavior of opportunistic criminals in an urban area. 
The idea is to plan the schedule of patrol units using the criminals' behavior learned as input. 
This work was extended \cite{zhang_keeping_2015} to use the algorithm \textit{Expectation Maximization}(EM) together with Bayesian networks to enhance the learning of opponent agents behavior (e.g.,\ opportunistic criminals in urban areas). Opponents behavior learning is also explored using Markov chains models with Monte Carlo methods to support \ac{MAS} planning \cite{panella_interactive_2017}. 
\ac{MAS} planning is alternated with opponents behavior learning to feed the plans generated. 
None of these approaches learn new cooperative multiagent plans. They use learned information about opponent behaviors to support the classical multiagent planning process.

A model for concurrent planning in a \ac{MAS} was presented using two learning approaches: Monte Carlo and \ac{LfD} \cite{yu_multiagent_2015}. The authors validated this work in a domain of static manipulator robots which cooperate in a \ac{MAS} to assemble an object. The nature of data in this environment is different from the data in mobile robots \ac{MRS} in a stochastic, real-time, partially observable environment (e.g.,\ robotic soccer). Nothing was mentioned about semantic equivalence (or any similar issue) when several domain experts provide slightly different recommendations with similar semantic meaning for domain experts. 
The interdependence of agents behavior in a \ac{MAS} was explored using the algorithm  Q-Learning and distributed Bayesian networks to model the planning of supply chains in a product global sales market \cite{zhou_combined_2015}. 
Another approach \cite{micalizio_explaining_2016} investigates fault (and their causes) detection in \ac{MAS} plans. The fault detection is associated with agents actions interdependence. 
Bayesian inference was used to diagnose the \ac{MAS} faults and their causes. None of these solutions provides new coordinated plans. 
Both works with learning or reasoning about behaviors interdependence which can be useful to deal with abort conditions in a setplay, for example. 
Synergy Graphs were presented for real-time training of coordination between agents in a \ac{MAS} \cite{liemhetcharat_allocating_2017}. The work used the well-known multi-armed bandit problem for validation. The authors did not mention if the learning process is restricted to the coordination between agents or if the process learns full coordinated plans with all their steps and agents' behaviors. In any case, no demonstrations from domain experts were used. 

The tools described in Section \ref{sec:tools} for setplays designing does not try to learn new setplays. 
They are a useful support for setplays designers, but no domain expert knowledge is directly extracted from demonstration based on real situations \cite{mota_co-ordination_2010}\cite{reis_playmaker:_2010}. Automatic analysis of match logs was used to generalize a setplay in a simulated robotic soccer environment from a sequence of successful events \cite{almeida_automatic_2013}. A sequence of behaviors derived from coordinated positioning is formalized as a plan and incorporated to the team setplays library for future use. The knowledge used emerges from the agents natural interaction. Agents do not use any domain expert knowledge for coordination.

A promising approach based on \ac{LfD} consists of hierarchical task decomposition \cite{freelan_towards_2014}. The goal is to split complex behaviors into simpler tasks. The authors presented a multiagent supervised learning system named \ac{HiTAB}.The latter represents agents behaviors as \acp{DFA}. Each state in a \ac{DFA} can represent a single atomic behavior or even a new \ac{DFA} at a lower level. Each level in the hierarchy represents a level of abstraction. Following the \ac{LfD} directive, the goal is to learn collaborative behaviors from scratch only using a human demonstrator to teach all the robot behaviors. The approach was validated in the RoboCup Humanoid Soccer League in RoboCup 2011 and 2012. One robot had all its previous programmed behavior deleted a few days before the RoboCup competitions and was trained by a human demonstrator using \ac{HiTAB} just a few days before the competition. The robot succeeded in participating in the \ac{MRS} performing the basic skills learned. In another experiment, the authors used an existing basic skills library to enable setplays learning. In this assay, two robots were individually trained, and then the two individual behaviors were joined into a higher level abstraction to form a setplay. The robots have no communication with each other. It was a simple pass and score setplay in the humanoid robotic soccer. One obvious limitation of this experience is the low number of robots ($2$) involved. For larger \ac{MRS} communication, synchronization and coordination must be present. This work did not mention anything related to semantic equivalence or treatment to inherent imprecision in the knowledge provided by several human demonstrators. 

Multiagent Q-Learning was used to learn a transition function for multi-flow setplays \cite{fabro_using_2014}. In multi-flow setplays, each state can lead to more than one following state, depending on the transition conditions. In this work, these transition conditions were generalized using reinforcement learning. Yet this proposal does not presents how agents can learn a complete setplay. A framework to enable many domain specialists to recommend new behaviors for agents defined the concept of \textit{crowdsourcing} \cite{moradi_learning_2016}. \textit{Crowdsourcing} means using the demonstrations from da very large number of domain experts to feed a dataset for learning. The authors used this approach for learning only individual agents control policy. The proposal does not present how agents could learn coordinated plans. The authors did not mention the semantic equivalence issue and any possible solution for it. 

A complete solution for decision making in a robotic soccer environment consists in aggregating the training dataset instances and the selection of adequate strategies using information from the environment, even if they are incomplete or imprecise \cite{shi_adaptive_2018}. 
This work combines Support Vector Machine (SVM), decision tree and reinforcement learning to generalize a complete control policy to the \ac{MRS}. Although, this approach does not say how the team can learn coordinated plans (e.g.,\ setplays). 
The solution learns decision-making policies for individual agents, and collaborative behavior is expected to emerge from the interaction of these individual policies. This approach for collaborative behavior is different from using setplays.  Another work presents a solution to transfer knowledge from one domain to another using Case-Based Reasoning (CBR) \cite{bianchi_heuristically_2017}. Although this proposal works in the domain abstraction level, it does not say how agents can learn cooperative behaviors or coordinated plans. It can be useful to transfer setplays learned in a domain (e.g.,\ RoboCup 3D Soccer Simulation) to another context (e.g.,\ RoboCup Humanoid Soccer League).

Many types of research try to treat the datasets to turn more adequate for the machine learning process. One of these works uses genetic algorithms to evaluate the similarity in the instances of groups generated by \ac{FCM} \cite{yang_hybrid_2012}. The main focus is on dataset size reduction. The authors did not mentioned semantic equivalence treatment or any similar concept.  

Another work describes a clustering solution to deal with datasets with incomplete information \cite{shao_clustering_2013}. Although incomplete information represents an important issue, it is different from the semantic equivalence problem. A promising solution uses a selection of features  representing symbolic objects \cite{ziani_feature_2014}. The idea presents some similarity with the solution presented in the current work because each symbolic object can represent a class of objects. There is no reference to any issue similar to the semantic equivalence. There was also no validation using any dataset related to \ac{MAS} coordinated plans or \ac{LfD}. 

Different works deals with optimization or preprocessing of very large datasets \cite{hieu_cell-mst-based_2015}\cite{nguyen_fast_2015}\cite{sardar_evaluation_2017}. Although it is not the main focus of the current work, we probably need to deal with large datasets if we use recommendations from many domain experts. However, none of these approaches deal with the semantic equivalence issue. 

Another work presents a method to reclassify the instances of a cluster generated by \ac{FCM} to minimize misclassification problems in these classes. The results of this work may complement the results from the current work. The single reclassification use cannot solve the semantic equivalence issue.

As far as we know, none of the state-of-the-art researches  deals with semantic equivalence between coordinated plans (e.g.,\ setplays) in a \ac{MAS}. Although some recent work use \ac{LfD} directives, none of them can learn a complete setplay from demonstration, considering the imprecision present in the domain-level knowledge. 
We can say that Section \ref{sec:dataset} presents a novel solution to organize setplays demonstrations to support  learning setplays in a \ac{MRS} dealing with the semantic equivalence issue.  

\section{A Dataset Schema for Learning Setplays}\label{sec:dataset}

SPlanner generates a setplay as a large text file using the S-expression syntax. After  collecting several setplays generated from recommendations from one or more domain experts, we can extract the relevant features representing each setplay. These features form a dataset schema which is used to generalize the instances in meaningful setplays in the domain-level knowledge.

A clustering solution based on \ac{FCM} is used to solve the semantic equivalence issue. The choice of \ac{FCM} is due to the inherent imprecision in the classification of a setplay. For example, one setplay can be classified as a triangulation, an offensive setplay, and a defensive setplay depending on the positioning of the players and the behaviors performed in the steps. It is more natural to consider a fuzzy clustering than strictly grouping these instances. 

A dataset schema was defined for clustering the setplays extracted from S-expression file analysis. Table \ref{tab:setplays-step} lists the features necessary to represent a setplay in the dataset.

\begin{longtable}{c|p{5cm}|p{3cm}}
     \caption{Dataset features representing setplays extracted from SPlanner.}
    \label{tab:setplays-dataset} \\
     \multicolumn{1}{c|}{\textbf{Feature}} & \multicolumn{1}{c|}{\textbf{Description}} & \multicolumn{1}{c}{\textbf{Data type}} \\ \hline \hline
    \endfirsthead

    \multicolumn{3}{c}%
    {{\bfseries \tablename\ \thetable{} -- continues from previous page}} \\
    \multicolumn{1}{c|}{\textbf{Feature}} &
    \multicolumn{1}{c|}{\textbf{Description}} &
    \multicolumn{1}{c}{\textbf{Data type}} \\ \hline \hline 
    \endhead

    \hline \multicolumn{3}{r}{{Continues in the next page}} \\ 
    \endfoot
    \hline \hline
    \endlastfoot

         ourPlayersNumber &  number of \ac{BahiaRT} players participating in the setplay  & Integer \\
          \hline
         theirPlayersNumber & number of opponent players participating in the setplay  & Integer \\
         \hline
         abortCondition & Boolean expression representing the condition to abort the setplay & Binary tree representing the parsed Boolean expression \\
         \hline
         Steps          & total number of steps composing the setplay & Integer \\
         \hline
         stepsList   & list of steps composing the setplay & Vector of Step. Each step is a composed structure presented in Table \ref{tab:setplays-step}.  \\
    \end{longtable}

\begin{figure}[htb]
    \centering
    \includegraphics[width=\textwidth]{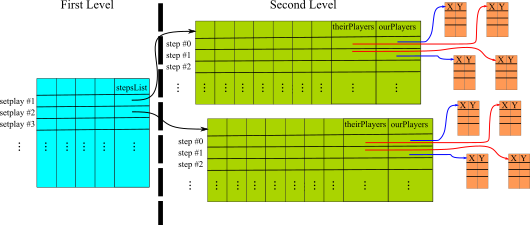}
    \caption{Schematic representation of the dataset schema. The first level contains the features from Table \ref{tab:setplays-dataset} that identify the setplays. The second level contains the features from Table \ref{tab:setplays-step} that describe the steps of each setplay.}
    \label{fig:dataset-schema}
\end{figure}

The last feature(stepsList) in  Table \ref{tab:setplays-dataset} is a list of complex structures composed of several features. A secondary dataset described in Table \ref{tab:setplays-step} details the stepsList . Figure \ref{fig:dataset-schema} illustrates a schematic view of the proposed dataset schema. The schema is divided into two levels. At the first level, there are features that identify the setplays. Each instance at this level represents a different setplay recommended by demonstrators. The second level describes the steps within each setplay. In the extraction phase, the S-expression files are read and transformed into the structure illustrated in Figure \ref{fig:dataset-schema}.

\begin{longtable}{c|p{4.5cm}|p{2.5cm}}
     \caption{Features that composes one step.}
    \label{tab:setplays-step} \\
     \multicolumn{1}{c|}{\textbf{Feature}} & \multicolumn{1}{c|}{\textbf{Description}} & \multicolumn{1}{c}{\textbf{Data type}} \\ \hline \hline
    \endfirsthead

    \multicolumn{3}{c}%
    {{\bfseries \tablename\ \thetable{} -- continues from previous page}} \\
    \multicolumn{1}{c|}{\textbf{Feature}} &
    \multicolumn{1}{c|}{\textbf{Description}} &
    \multicolumn{1}{c}{\textbf{Data type}} \\ \hline \hline 
    \endhead

    \hline \multicolumn{3}{r}{{Continues in next page}} \\ 
    \endfoot
    \hline \hline
    \endlastfoot

         ourPlayersInStep &  Number of \ac{BahiaRT} players  participating in this step. This number must be  smaller than or equal to ourPlayersNumber & Integer \\
          \hline
         theirPlayersInStep & Number of opponent participating in this step. This number must be smaller than or equal to theirPlayersNumber & Integer \\
         \hline
         waitTime & minimum time to wait before a transition from a current step to another can be conducted & Double \\
         \hline
         abortTime          & maximum time to finish the current step. & Double \\         \hline
         ourPlayersList   & List of \ac{BahiaRT} players participating in this step. & Vector of Player $P_B$. Each Player $p_B \in P_B$ is represented by a Cartesian coordinates pair $(x_{p_B},y_{p_B})$ . \\
           \hline
         theirPlayersList   & List of opponent players participating in this step. & Vector of Player $P_O$. Each Player $p_O \in P_O$ is represented by a Cartesian coordinates pair $(x_{p_O},y_{p_O})$ \\
            \hline
         nextStep   & identification of the next step after a transition condition is met. & Integer \\
         \hline
         condition   & Boolean expression determining the transition condition to the next step & Binary tree representing the parsed Boolean expression. \\
          \hline
         behaviorsList   & list of behaviors executed by each player $p_B \in P_B$. & Vector of strings. \\
    \end{longtable}

To perform the clustering of the dataset organized in this schema, we use the \ac{FCM} in two stages. At the first stage, only the features at the first level of the dataset schema are considered, except for the \textit{stepsList} feature. The first stage generates a new dataset named \textit{Grouped dataset 1}. 
This dataset contains a set of clusters of semantic equivalent setplays, considering only the similarity between the first four features at the first level of the dataset schema. 
The idea is that setplays with a similar number of steps, abort conditions and number of players can be grouped under the same class. 
As we are using a Fuzzy approach, one setplay can be part of more than one class with different degrees of membership.

At the second stage, we execute many instances of \ac{FCM}, one for each class in the Grouped dataset 1. The idea is to consider each cluster as the base dataset for each \ac{FCM} execution. As a result, Grouped dataset 1 contains clusters with many sub-clusters. The set of all sub-clusters generated in this second stage forms the \textit{Grouped dataset 2}. At this second stage, the algorithm performs a fine-grained evaluation of semantic evaluation. Now the features at the second level of the dataset schema are considered. Setplays with similar steps descriptions can be grouped under the same sub-cluster. 

\begin{figure}[htb]
    \centering
    \includegraphics[width=.8\textwidth]{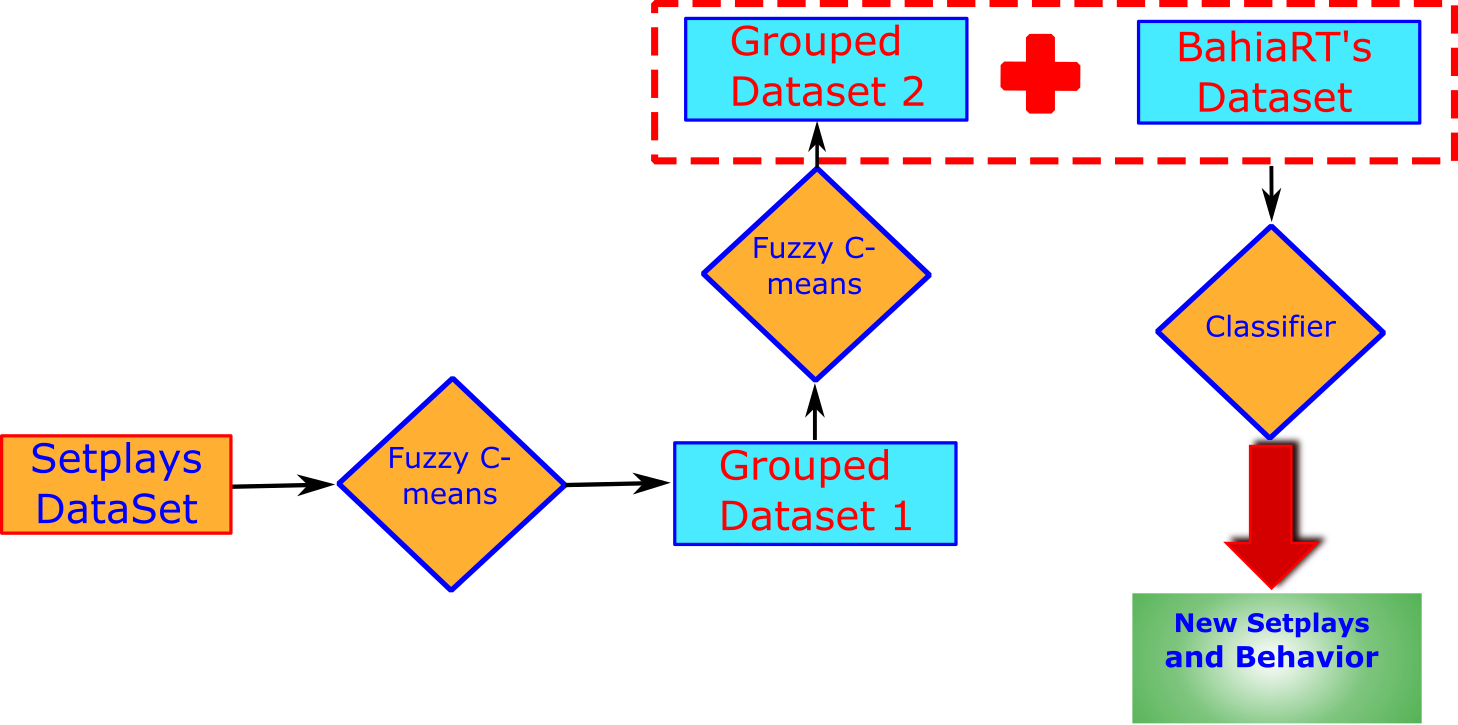}
    \caption{Schematic representation of the \ac{FCM} two-stage solution to identify classes of semantic equivalent setplays.}
    \label{fig:fcm-two-stages}
\end{figure}

The motivation to split the solution into two stages was to reduce the complexity of evaluating semantic equivalence. As the concept of semantic equivalence is itself ambiguous, we provide a bias to the clustering solution. Initially, we use only the features in the first level of the dataset schema to foster that setplays with important differences in the number of players or steps not to be grouped in the same cluster of semantic equivalence. Also, setplays with considerable differences in the abort condition are not grouped in the same cluster. Of course, as in fuzzy approach, obtaining some setplays with differences in these features from the others in the same cluster is possible, but presenting a lower degree of membership. Figure \ref{fig:fcm-two-stages} illustrates the complete solution for setplay learning.

When the \ac{FCM} outputs the Grouped dataset 2, it is joined with the previously validated \ac{BahiaRT} world model dataset \cite{simoes_towards_2018}  describing the environment situations whereby each setplay can be fired. The classifier will learn a decision-making policy to decide if a setplay will be fired or not. If the policy chooses a setplay to be fired, the choice is for a group of semantic equivalent setplays. Then, all the setplays in this group and their degrees of membership will be sent to the current \ac{FSF} setplay manager algorithm to choose which instance of the chosen class will effectively be fired. 

We use the classical \ac{FCM} algorithm~\cite{bezdek_pattern_1981}\cite{nayak_fuzzy_2015} and the unique pending definition to apply it to our dataset schema is a norm to estimate the distance between two instances in the dataset. The Euclidean distance is commonly used as a norm to estimate the distance between two instances for scalar features. The proposed dataset schema contains some non-scalar data types. The next subsection defines a norm to estimate the distance between two instances, considering the data types heterogeneity present in our dataset schema.

\subsection{A norm to estimate the distance between two instances in the dataset schema}

Consider a set of instances $X=\{x_1, x_2, \ldots, x_n\}$; where $n$ is the number of instances in the dataset $X$. Each instance $x_i; \forall i=1,\ldots,n$ is a vector representing the $5$ features described in the Table \ref{tab:setplays-dataset}. So, $x_i=\{x_{i,1}, x_{i,2}, \ldots, x_{i,5}\}$ represents the values of the features in the Table \ref{tab:setplays-dataset} for the instance $x_i$. 

Given two instances $x_i$ and $x_j; \forall i,j=1,\ldots n; i \ne j$, the distance between $x_i$ and $x_j$ is defined by:

\begin{equation}\label{eq:norma-setplay}
\resizebox{\textwidth}{!}%
{
    $d\left(x_i,x_j\right)=d\left(x_j,x_i\right)=\sqrt{\left(x_{j,1} - x_{i,1}\right)^2+\left(x_{j,2} - x_{i,2}\right)^2+\norm*{x_{j,3} - x_{i,3}}^2+\left(x_{j,4} - x_{i,4}\right)^2}$
}
\end{equation}

Notice that features $x_{i,5}$ and $x_{j,5}$ are not considered at the first stage of the \ac{FCM} solution. For this reason their values are not used to calculate the distance between $x_i$ and $x_j$. $x_{i,5}$ and $x_{j,5}$ are \textit{stepsLists} and their values will be considered in the second stage of the \ac{FCM} execution. 

Features $x_{i,3}$ and $x_{j,3}$ need a special norm because they are non-scalar data types. $x_{i,3}$ and $x_{j,3}$ are binary trees representing Boolean expressions. All the Boolean expressions in this dataset are automatically generated from SPlanner from a finite set of conditions. The structure of the  binary trees generated presents a few variations. Therefore,  the norm for $x_{i,3}$ and $x_{j,3}$ can be defined as:

\begin{equation}\label{eq:norma-arvore}
    \norm*{x_{j,3} - x_{i,3}}=DiffNode\left(x_{i,3},x_{j,3}\right)
\end{equation}

\noindent where $DiffNode\left(x_{i,3},x_{j,3}\right)$ is the number of different nodes in the trees represented in  $x_{i,3}$ and $x_{j,3}$. The remaining features are scalar and the norm is calculated in a similar fashion to the Euclidean distance. 

The norms defined in \eqref{eq:norma-setplay} and \eqref{eq:norma-arvore} are used to compute the objective function and iterative update of the partition matrix in the \ac{FCM} algorithm.

At the second stage of the \ac{FCM} execution, the instances in the Grouped dataset 1 ($GD^{(1)}$) will be considered. $GD^{(1)} = \{\psi_1, \ldots, \psi_{C^{(1)}}\}$, where $C^{(1)}$ is the number of clusters generated in the first stage. Each $\psi_a; a=1,\ldots,C^{(1)}$ is a set of instances considered as the initial dataset for each \ac{FCM} execution in the second stage.  The dataset is defined by $Y_{a,k}=\{y_{a,k,1}, y_{a,k,2}, \ldots, y_{a,k,m_k}\}$; where $m_k$ is the number of steps of the $k^{th}$ setplay in the cluster $\psi_a$ derived at the first stage, and used as a dataset at each \ac{FCM} execution at the second stage.  $k=1,\ldots,G$; where $G$ is the total number of setplays belonging to the current dataset.

Given two setplays in the current cluster, $Y_{a,k}$ e $Y_{a,l}; \forall k=1,\ldots,m_k; l=1,\ldots,m_l$; where $m_k$ e $m_l$ are the number of steps of $Y_{a,k}$ e $Y_{a,l}$. The distance between $Y_{a,k}$ e $Y_{a,l}$ is defined as:
\begin{equation}\label{eq:norma-lista-steps}
    d\left(Y_{a,k},Y_{a,l}\right)^{(2)}=d\left(Y_{a,l},Y_{a,k}\right)^{(2)}=d\left(Y_{a,l},Y_{a,k}\right)^{(1)}+ \sum_{z=1}^{\min\left(m_k,m_l\right)}  \norm*{y_{a,l,z} - y_{a,k,z}}
\end{equation}

 \noindent where $d\left(Y_{a,l},Y_{a,k}\right)^{(1)}$ and $d\left(Y_{a,l},Y_{a,k}\right)^{(2)}$ are the distances between setplays $Y_{a,l}$ and $Y_{a,k}$ at the first and second stage respectively in the dataset $\psi_a$. $d\left(Y_{a,l},Y_{a,k}\right)^{(1)}$ is calculated as defined in \eqref{eq:norma-setplay}.$z=1,\ldots,\min\left(m_k,m_l\right)$ is the step number of setplays $Y_{a,l}$ and $Y_{a,k}$.

Each step in both setplays is compared, computing the distance between them. Each step $y_{a,k,z}$ or $y_{a,l,z}$ is a vector of features described in Table \ref{tab:setplays-step}. $y_{a,k,z}=\{p_{a,k,z,1},\ldots, p_{a,k,z,9}\}$ and $y_{a,l,z}=\{p_{a,l,z,1},\ldots, p_{a,l,z,9}\}$ represent the $9$ features representing each step. The distance between two steps is defined by:

\begin{dmath}
    \norm*{y_{a,l,z} - y_{a,k,z}} = \sqrt{\left(p_{a,l,z,7}-p_{a,k,z,7}\right)^2 + \sum_{i=5,6,8,9}\norm{p_{a,l,z,i} - p_{a,k,z,i}}^2 + \sum_{j=1}^{4} \left(p_{a,l,z,j} - p_{a,k,z,j}\right)^2}
\end{dmath}

Following the same criteria used at the first stage, the features with scalar data types (e.g.\ $p_{a,l,z,i} and p_{a,k,z,i}; i=1,2,3,4,7$) use a calculation similar to the one used in Euclidean distance. The remaining features use specific norms according to their data type. $p_{a,l,z,8}$ and $p_{a,k,z,8}$ are binary trees with Boolean expressions. Their norm is defined using the same method as the one used in \eqref{eq:norma-arvore}. 

Features $p_{a,l,z,9}$ and $p_{a,k,z,9}$ are vectors of strings representing the behaviors of each player at steps $y_{a,k,z}$ and $y_{a,l,z}$. The norm compares the two vectors and counts the number of different behaviors. The norm for $p_{a,l,z,9}$ and $p_{a,k,z,9}$ is defined as:

\begin{equation}
    \label{eq:strings-norm}
    \begin{split}
        \norm*{p_{a,l,z,9} - p_{a,k,z,9}} & = \sqrt{\sum_{i=1}^{\max (B_{a,l,z}, B_{a,k,z})}\Delta_i};%
        \\
        \Delta_i & = \begin{cases}%
                        1, if\ \beta_{a,l,z,i}\neq\beta_{a,k,z,i}\\%
                        0, otherwise%
                    \end{cases}%
    \end{split}
\end{equation}

\noindent where $B_{a,l,z}$ and $B_{a,k,z}$ are the number of behaviors in vectors $p_{a,l,z,9}$ and $p_{a,k,z,9}$ respectively; $\Delta_i$ is an accumulator; $\beta_{a,l,z,i}$ and $\beta_{a,k,z,i}$ are individual strings representing behaviors in vectors $p_{a,l,z,9}$ and $p_{a,k,z,9}$.

The remaining features $p_{a,l,z,5}$,  $p_{a,k,z,5}$, $p_{a,l,z,6}$ and $p_{a,k,z,6}$ are player lists. Each player is defined by a Cartesian coordinated by the ordered pair $\left(x, y\right)$, where  $x$ and $y$ are the regions where each player is supposed to be at the beginning of each step. The Euclidean distance can define the distance between two players. The sum of all distances between pairs of players defines the norm of these lists of players. 

The \ac{FCM} algorithm can be executed both at the first and second stages using the defined norms to generate a set of clusters and sub-clusters representing the semantic equivalence concept. The next Section presents the assessment of this solution.

\section{Assessment}\label{sec:assessment}

The dataset schema proposed to organize setplays into clusters is based on fuzzy clustering as stated in previous sections. In this section, we assess if the  schema presented meets the requirements to organize the full dataset of setplays into a number of groups. One of the goals when using \ac{FCM} is to define the best number of clusters to adequately represent the system of interest. A widely known and simple scheme for defining the number of clusters consists in executing the fuzzy clustering algorithm several times for different numbers of clusters and then selecting the particular number of clusters that provides the best result according to a specific criterion~\cite{babuska_fuzzy_2012}\cite{hoppner_fuzzy_1999}.

Many \acp{CVI} were proposed and analyzed in recent work~\cite{eustaquio_monotonic_2018}\cite{eustaquio_fuzzy_2018}. These \acp{CVI} are used to assess the quality of data organization into clusters. \acp{CVI} are popular measures to assess the number of clusters used in a particular data organization. We use \ac{FS} to verify if our proposed data schema can provide a good organization for setplays and represent the concept of semantic equivalence in an adequate number of clusters. \ac{FS} non-monotonic bias, good scalability to large datasets, and low computation costs are the main reasons for our choice~\cite{campello_fuzzy_2006}. 

For the assessment, we used the \ac{FCM} algorithm to split a sample dataset into clusters. The dataset was generated using SPlanner. The dataset was composed of $18$ setplays. To provide diversity, the setplays was created in four different play-modes: play on, goal kick, kick in and kick-off. At least four different setplays were created for each play mode. The number of players in each setplay presents a variation from $1$ to $8$ and the number of steps varies from $2$ to $6$. Hence, the dataset is composed both by simple and complex setplays. The next subsection briefly describes the \ac{FS} \ac{CVI}.  

\subsection{\acf{FS}}\label{sec:silhueta}

Consider the fuzzy partition matrix $P=\left[\mu_{i,j}\right]_{CxN}$; where $C$ is the number of clusters used in the \ac{FCM} algorithm, $N$ is the number of objects to be organized into clusters, $\mu_{i,j}$ is the membership degree of object $j$ to cluster $i$, $i=1, \ldots, C$, $j=1, \ldots, N$. The \ac{FS} is defined by~\cite{campello_fuzzy_2006}:

\begin{equation}\label{eq:silhuetafuzzy}
    FS=\frac{\sum_{j=1}^N \left(\mu_{p,j} - \mu_{q,j}\right)^{\alpha} s_j}{\sum_{j=1}^N \left(\mu_{p,j} - \mu_{q,j}\right)^{\alpha}},%
\end{equation}

\noindent where $\mu_{p,j}$ and $\mu_{q,j}$ are the first and second largest elements of the $jth$ column of the fuzzy partition matrix, respectively, and $\alpha$ is a user-defined weighting coefficient. $s_j$ is the silhouette of object $j$ defined as follows:

\begin{equation}\label{eq:silhueta}
    s_j=\frac{b_{p,j} - a_{p,j}}{\max\left({a_{p,j},b_{p,j}}\right)},%
\end{equation}

\noindent where $a_{p,j}$ is the average distance of object $j$ to all other objects \textit{belonging} to cluster $p$. The distance is calculated using the norms defined in Section \ref{sec:dataset}. \textit{Belonging} to cluster $p$ means that the membership of the $jth$ object to the $pth$ fuzzy cluster, $\mu_{p,j}$, is higher than the membership of this object to any other fuzzy cluster, i.e., $\mu_{p,j}>\mu_{q,j} \forall q \in \left\{1,\ldots,c\right\}, q \neq p$. Let $d_{q,j}$ be the average distance object $j$ to all objects \textit{belonging} to another cluster $q$, $q \neq p$. $b_{p,j}$ is the minimum $d_{q,j}$ computed over $q=1, \ldots, c$, $q \neq p$. Exponent $\alpha$ is an optional user-defined parameter ($\alpha=1$, by default). When $\alpha$ approaches zero, the $FS$ measure defined in \eqref{eq:silhuetafuzzy} approaches the \ac{CS} measure which serves as a basis to define \ac{FS} \cite{campello_fuzzy_2006}.  \ac{CS} is the \textit{crisp} counterpart of \ac{FS} and is used for \textit{crisp} clustering algorithms applied to hard datasets whereby no overlap between clusters is present. Conversely, increasing $\alpha$ moves \ac{FS} away from \ac{CS} by diminishing the relative importance of data objects in overlapping areas. Accordingly, increasing $\alpha$ tends to stress the effect of revealing smaller regions with higher data densities (sub-clusters), if they exist. Such an effect can be particularly useful, for example, when dealing with data sets contaminated by noise. Bearing
this in mind, exponent $\alpha$ can be seen as an additional tool for exploratory data analysis, as is the case with fuzzifier exponent ($m$) of the \ac{FCM} algorithm. From another perspective, the \ac{FS} with exponent $\alpha$ can be seen as a family of parameterized \acp{CVI}, rather than a single measure with a coefficient that must be adjusted to a specific problem in hand. We here used four combinations of $m$ and $\alpha$, as follows: $(m=1.5,\alpha=1); (m=1.5, \alpha=2); (m=2, \alpha=1); (m=2, \alpha=2)$.

\ac{FS} is a maximization \ac{CVI}. Hence, the higher the value of \ac{FS} for a particular value of $c$, the better this specific clustering is. The goal when using \ac{FS} to evaluate the clusters setup generated by \ac{FCM} is to find the value of $c$ that maximizes \ac{FS}.

\subsection{Experimental setup}\label{sec:experiment}

We adapted the standard \ac{FCM} implementation~\cite{bezdek_pattern_1981} to use the norms defined in Section \ref{sec:dataset}. The algorithm was also developed to split the clustering process into two stages as described. At the first stage the feature \textit{stepsList} of the setplays objects is ignored. When the centroids of each cluster are defined, the \textit{abortCond} feature is assigned the binary tree from the \textit{abortCond} property of the instance with a higher membership value in the fuzzy partition matrix. 

We populated our dataset with setplays assigned to four different setplays categories based on their current game play-mode. We  thus considered $C^{(1)}=2, \ldots, 8$ to run the \ac{FCM} and to find different organizations for the dataset. Our experiment consists of a $18$-instance dataset created from $4$ different group of setplays. So our expectation is to get something near $C^{(1)}=4$ clusters of setplays. To assess if this expectation is confirmed we decided to vary the value of $C^{(1)}$ up to $2\times4=8$ groups.  \ac{FS} where used to assess each setup with different values of $c$. For each value of $c$, we ran $10$ instances of the \ac{FCM} algorithm initializing with a random prototype of clustering. The higher value of \ac{FS} among the $10$ instances was considered the representative value for that $c$ particular instance.

After running the experiment for the first stage we found the optimal value $C^{(1)^*}$ for $C^{(1)}$. We used a cluster instance for $C^{(1)}=C^{(1)^*}$ to start the second stage. 

To define the membership of dataset instances for each of the $C^{(1)^*}$ clusters, consider $\Delta\mu_i =\left(\max \mu_{i,j}\right)- \gamma$ ; $j=1, \ldots, N; i=1, \ldots, C^{(1)^*}$; where $\gamma \in \left[0,1\right]$  is a constant used to define how flexible the membership condition was. Greater values of $\gamma$ tend to give more flexibility to the membership condition, i.e., instance of setplays with membership degrees far from the best-valued instance were also considered as a member of that particular cluster. When $\gamma$ tends to $1$, all the instances of setplays in the dataset were considered  members of the $ith$ cluster. When $\gamma$ tended to $0$ only the best-valued instance was considered a member of the $ith$ cluster. The clustering setup degenerated to a set of singletons, i.e., groups with only one instance.

We defined that the $jth$ instance was a member of the $ith$ cluster if $u_{i,j} > \Delta\mu_i$. In this work, we used $\gamma=0.5$ for all the experiments.

\subsection{Results}

\begin{figure}[h]%
    \centering
    \subfigure[FIGBOTCAP][$m=1.5~\alpha=1$ ]{\label{fig:fs-level1-a}\includegraphics[width=0.47\textwidth]{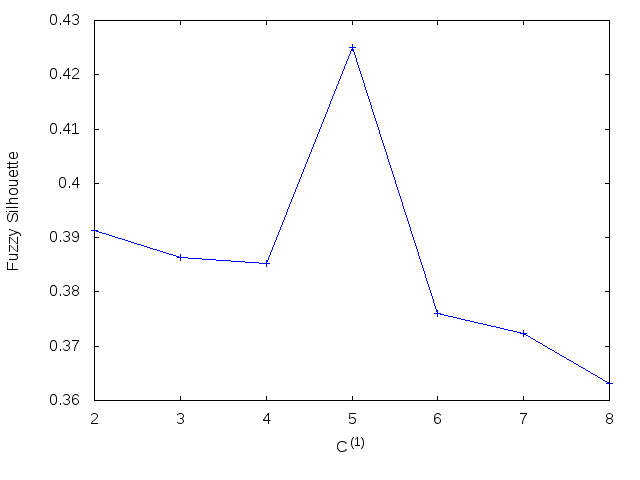}}
    \qquad
    \subfigure[FIGBOTCAP][$m=1.5~\alpha=2$ ]{\label{fig:fs-level1-b}\includegraphics[width=0.47\textwidth]{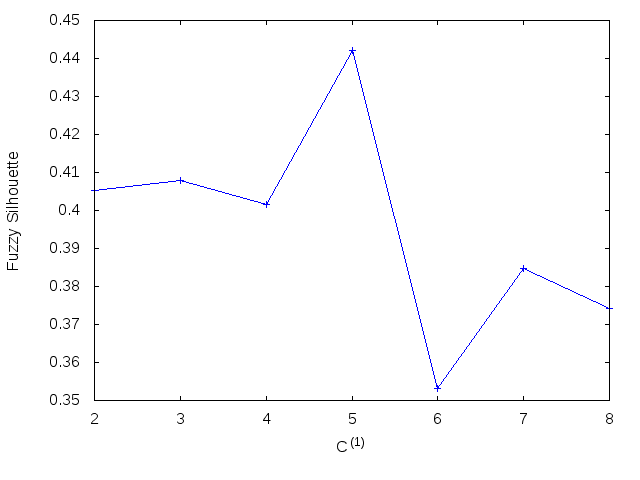}}
    \newline
    \subfigure[FIGBOTCAP][$m=2~\alpha=1$ ]{\label{fig:fs-level1-c}\includegraphics[width=0.47\textwidth]{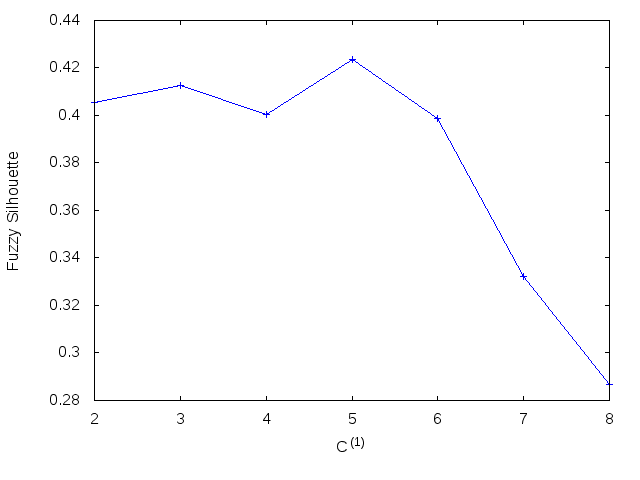}}
    \qquad
    \subfigure[FIGBOTCAP][$m=2~\alpha=2$ ]{\label{fig:fs-level1-d}\includegraphics[width=0.47\textwidth]{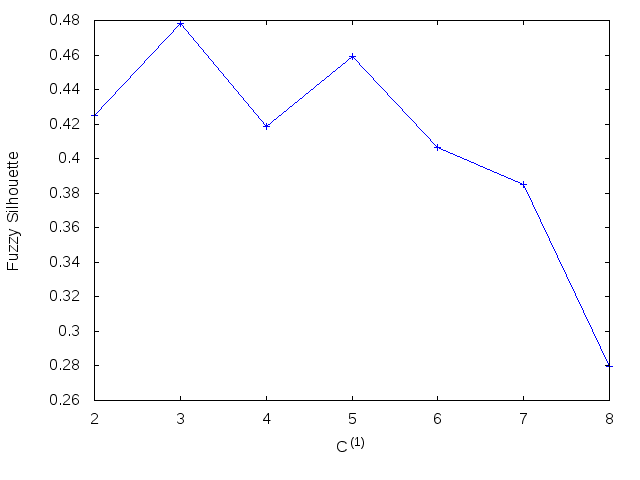}}
    \caption{Results of \ac{FS} for $C^{(1)}=2, \ldots, 8$ getting the best solution after $10$ executions for each value of $C^{(1)}$.} 
    \label{fig:results-lvl1}
\end{figure}

After the first stage we could assess the first level of our dataset schema. Figure \ref{fig:results-lvl1} shows the results. In most combinations of $m$ and $\alpha$, the best configuration was obtained when $C^{(1)}=5$. This result was expected, since we created $4$ different groups of setplays, one for each of four different play-mode situations (e.g., play-on, kick-in, goal-kick and kick-off). Thus, an optimal value for $C(1)$ near the original number of groups ($4$) was expected. 

In Figure \ref{fig:results-lvl1}\subref{fig:fs-level1-d}, the best value for $C^{(1)}$ is $3$, but its \ac{FS} value is very close to the measurement for $C^{(1)}=5$. In Figure \ref{fig:results-lvl1}\subref{fig:fs-level1-c}, the values for \ac{FS} are closer to the values exhibited in Figure \ref{fig:results-lvl1}\subref{fig:fs-level1-d} than those in the other two configurations. It is a clear result of the influence of the fuzzifier exponent ($m=2$). Higher values of $m$ give a bias to the algorithm to explore the fuzzy nature of the dataset assigning more relevance to overlapping features. In Figures \ref{fig:results-lvl1}\subref{fig:fs-level1-a} and \ref{fig:results-lvl1}\subref{fig:fs-level1-b}, when the value of $m=1.5$, the clusters exhibit less overlapping and the \ac{FS} for $C^{(1)}=5$ presents a greater difference to the other clusters' setup than in the other two figures. 

When visualizing the membership of the configuration with $C^{(1)}=5$, a clear semantic equivalence is observed in each cluster. Simpler setplays are in the same cluster and more complex setplays are grouped in other clusters. In each cluster, the difference between the number of teammate players, number of opponnent players or number of steps are always smaller than $2$.  Most of the instances in the same cluster are related to the same play-mode situation, or similar tactical situations, e.g. offensive or defensive situations. The experiment  showed that, at a first level, it can represent the semantic equivalence. However, there are different situations within the same group, i.e., at a first glance, a kick-in setplay and a kick-off setplay in the same cluster do not seems the most adequate setup. This kind of situation is explained because in the first level, we did not consider the features related to players' positions and behaviors. The second stage of the \ac{FCM} takes these features into account.

\begin{figure}[h]%
    \centering
    \subfigure[FIGBOTCAP][$c^{(1)}=1$ ]{\label{fig:fs-level2-a}\includegraphics[width=0.47\textwidth]{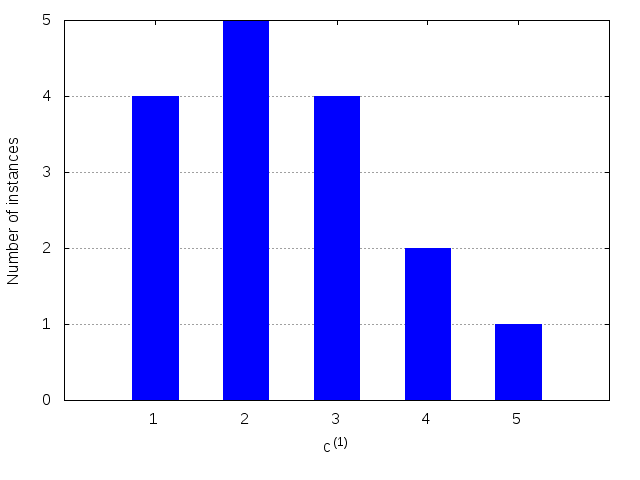}}
    \qquad
    \subfigure[FIGBOTCAP][$c^{(1)}=2$ ]{\label{fig:fs-level2-b}\includegraphics[width=0.47\textwidth]{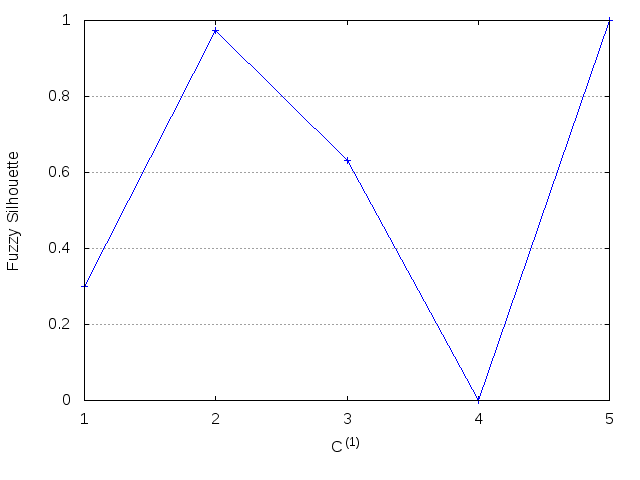}}
    \caption{Results of \ac{FS} for the second stage of \ac{FCM}. Figure \subref{fig:fs-level2-a} exhibits the number of instances of the dataset belonging to each cluster generated in the first stage. Figure \subref{fig:fs-level2-b} shows the \ac{FS} value for executing \ac{FCM} algorithm for each cluster of first level $c^{1}=1, \ldots, 5$ using $c^{(2)}=2$.} 
    \label{fig:results-lvl2}
\end{figure}

In the second stage, the execution of \ac{FCM} was held for each subset of setplays grouped into each of $C^(1)=5$ clusters defined in the first stage. Each cluster in the first stage is identified by $c^{(1)}=1,\ldots,C^(1)$. The number of clusters in the second stage is defined as $C^{(2)}$ and the \ac{FCM} algorithm was executed for  $C^{(2)} = 2$. The choice for using this unique value for maximum number of clusters is due to the low number of instances in each cluster generated in the first stage. If we use higher values for $C{(2)}$ the solution tends to degenerate to a $C=N$ setup with one cluster per instance. These unitary clusters are named \textit{singletons}. It is usual to consider the maximum value for $C^{(2)}=\sqrt{N}$, where $N$ is the number of instances in each the dataset used as input for \ac{FCM}. At the second stage, the \ac{FCM} was executed using $C^{(2)}=2, m=2, \alpha=1$. Figure \ref{fig:results-lvl2} depicts the configuration of clusters of the first level used as input for the \ac{FCM} algorithm at the second stage and the \ac{FS} index measure for the clusters generated at the second stage. For each $c(1) \in \left[1,C^{(1)}\right]$, the \ac{FCM} was executed $10$ times and the highest value of \ac{FS} is presented in Figure \ref{fig:results-lvl2}\subref{fig:fs-level2-b}.  

Figure \ref{fig:results-lvl2}\subref{fig:fs-level2-a} shows the number of instances in each cluster resulting from the first stage. The groups resulting from first stage present only one \textit{singleton}($c^{(1)}=5$ containing a single kick-off setplay with only two steps. It is the trivial kick-off setplay case in which the goalie just performs a long pass to one of the teammates in the middle of the field.  Singletons can not be divided into other \textit{sub-clusters}. To deal with this situation, we considered $FS=1$ for these cases. In Figure \ref{fig:results-lvl2}\subref{fig:fs-level2-b}, it is possible to see that $FS=1$ for the singleton $c^{(1)}=5$. The other four clusters were successfully split into $C^(2)=2$ \textit{sub-clusters} each. Two instances of setplays belong to cluster $c^{(1)}=4$. When this cluster was split into two \textit{sub-clusters} the value of \ac{FS} tends to $0$ because the two setplays in $c^{(1)}=4$ are too similar to each other. Both are kick-in setplays in close regions of the fields and the same number of steps. Thus, the distance between the players and the behaviors in the list of steps is short and the two setplays are too close in the dataset. This scenario derives a low value for \ac{FS}. Clusters $c^{(1)}=1, \ldots, 3$ present higher values of \ac{FS} when divided into  $C^(2)=2$ \textit{sub-clusters}. The resulting configuration shows semantic equivalent groups containing setplays that represent the same play-mode situation or tactical situation. The results allow concluding that we would split the $c^{(1)}=1, \ldots, 3$ into $C^(2)=2$ \textit{sub-clusters} but we should keep $c^{(1)}=4, \ldots, 5$ as a single cluster for the final configuration of the dataset.  

\section{Conclusions and Future Work}\label{sec:conclusion}

This work presents a novel schema to represent coordinated plans in a \ac{MRS}. The dataset includes non-scalar features such as Boolean conditions, lists of complex objects (i.e., list of behaviors, list of players, list of steps). The dataset schema is divided into two levels to deal with the complexity of these non-scalar data types. The main motivation to create this schema is to represent the semantic equivalence defined in Section \ref{sec:semantic-equivalence}. This work is part of a project that aims to enable domain experts to suggest new coordinated plans from real situations in which robots  act in an \ac{MRS}. Big datasets, formed by recommendations from several domain experts, tend to present a number of semantic equivalent instances. These instances may have different features values but the meaning in the abstract human knowledge level is the same. 

If learning approaches are applied to the rough dataset, the solution will present high computational costs and may present problems with over fitting due to a high number of semantic equivalent instances. Therefore, the main goal of this work is to propose a dataset schema to organize the original dataset into groups of semantic equivalent instances. The concept of semantic equivalence has no strict limit, because it is in the human abstract knowledge level. Thus, one instance can be semantic equivalent to instances of two different groups with different degrees of membership. For this reason, we considered the use of a fuzzy approach to organize the dataset.

The \ac{FCM} algorithm was used in a two-stage implementation to deal with the two-level structure of the proposed dataset schema. The \ac{FS} clustering validation index was used to assess the cluster configuration generated by the \ac{FCM} execution. Besides using an objective measure to assess the groups configuration, it is important to visually check which instance is assigned to each cluster. Since semantic equivalence is an abstract concept, the assessment should consider this aspect when evaluating if the cluster configuration makes sense. In our assessment, we used a dataset generated in our lab using the SPlanner tool. 

The results show an useful group configuration for future setplays learning experiments. The clusters contain semantic equivalents setplays and some hidden similarities were discovered due to the fuzzy aspect of the solution used to assess the dataset schema. The objective \ac{FS} measure shows that clusters in the second level can be representative of the semantic equivalence or not. Clusters at the first level with higher number of instances tend to generate more adequate \textit{sub-clusters} at the second level. It is also possible to use \ac{FS} to define if a group at the first level will be split into two or more groups at the second level or if it should be kept as a single group. 

Hence, we can conclude that the proposed dataset schema can be used as a base to organize a setplays dataset in an \ac{MRS} into semantic equivalent groups of instances. Using this contribution, it is possible to support the setplays learning from a demonstration project. As future work, we plan to use other cluster validity indexes to compare the measurements with the results of this work. Larger datasets built from different domain experts recommendations will be used to acknowledge the results presented. The results of these experiments using real databases will be used to assess the setplays learning from demonstration proposal. 

Although this work has applied all the assessment to the robotic soccer domain, any \ac{MRS} using coordinated can benefit from the results presented here. Conditions and list of behaviors and steps are common features in any coordinated plans for robots in an \ac{MRS}. 

\section*{Acknowledgements}
This is a pre-print of an article published in Journal of Intelligent \& Robotic Systems. The final authenticated version will be available online at: \url{https://doi.org/10.1007/s10846-019-01123-w}.

%
%


\bibliographystyle{spmpsci}      
\bibliography{references}   

\end{document}

%% file: acronimos.tex
\acrodef{MAS}{Multi-Agent System}
\acrodef{MRS}{Multi-Robot System}
\acrodef{KQML}{Knowledge Query and Manipulation Language}
\acrodef{KIF}{Knowledge Interchange Format}
\acrodef{LfD}{Learning from Demonstration}
\acrodef{FCM}{\textit{Fuzzy C-Means}}
\acrodef{DFA}{deterministic finite automaton}
\acrodef{FSF}{FCPortugal Setplay Framework}
\acrodef{BahiaRT}{Bahia Robotics Team}
\acrodef{AI}{Artificial Intelligence}
\acrodef{SPlanner}{Setplays Planner}
\acrodef{HiTAB}{\textit{Hierarchical Training of Agent Behaviors}}
\acrodef{CVI}{Cluster Validation Index}
\acrodef{FS}{Fuzzy Silhouette}
\acrodef{CS}{\textit{Crisp} Silhouette}

%% file: ms.bbl
\begin{thebibliography}{10}
\providecommand{\url}[1]{{#1}}
\providecommand{\urlprefix}{URL }
\expandafter\ifx\csname urlstyle\endcsname\relax
  \providecommand{\doi}[1]{DOI~\discretionary{}{}{}#1}\else
  \providecommand{\doi}{DOI~\discretionary{}{}{}\begingroup
  \urlstyle{rm}\Url}\fi

\bibitem{almeida_automatic_2013}
Almeida, F., Abreu, P.H., Lau, N., Reis, L.P.: An automatic approach to extract
  goal plans from soccer simulated matches.
\newblock Soft Computing \textbf{17}(5), 835--848 (2013).
\newblock \doi{10.1007/s00500-012-0952-z}.
\newblock \urlprefix\url{https://doi.org/10.1007/s00500-012-0952-z}

\bibitem{babuska_fuzzy_2012}
Babuška, R.: Fuzzy modeling for control, vol.~12.
\newblock Springer Science \& Business Media (2012)

\bibitem{bezdek_pattern_1981}
Bezdek, J.C.: Pattern {Recognition} with {Fuzzy} {Objective} {Function}
  {Algorithms}.
\newblock Kluwer Academic Publishers, Norwell, MA, USA (1981)

\bibitem{bianchi_heuristically_2017}
Bianchi, R.A.C., Santos, P.E., da~Silva, I.J., Celiberto, L.A., Lopez~de
  Mantaras, R.: Heuristically {Accelerated} {Reinforcement} {Learning} by
  {Means} of {Case}-{Based} {Reasoning} and {Transfer} {Learning}.
\newblock Journal of Intelligent \& Robotic Systems  (2017).
\newblock \doi{10.1007/s10846-017-0731-2}.
\newblock \urlprefix\url{https://doi.org/10.1007/s10846-017-0731-2}

\bibitem{breiman_classification_2017}
Breiman, L.: Classification and {Regression} {Trees}.
\newblock CRC Press, Boca Raton, FL, USA (2017).
\newblock \urlprefix\url{https://books.google.com.br/books?id=gLs6DwAAQBAJ}

\bibitem{campello_fuzzy_2006}
Campello, R., Hruschka, E.: A fuzzy extension of the silhouette width criterion
  for cluster analysis.
\newblock Fuzzy Sets and Systems \textbf{157}(21), 2858--2875 (2006).
\newblock \doi{10.1016/j.fss.2006.07.006}.
\newblock
  \urlprefix\url{https://www.scopus.com/inward/record.uri?eid=2-s2.0-33749142135&doi=10.1016%2fj.fss.2006.07.006&partnerID=40&md5=383e6afe20ad96ea4b9e8bfda1f14ebe}

\bibitem{dambrosio_scalable_2013}
D'Ambrosio, D.B., Stanley, K.O.: Scalable multiagent learning through indirect
  encoding of policy geometry.
\newblock Evolutionary Intelligence \textbf{6}(1), 1--26 (2013).
\newblock \doi{10.1007/s12065-012-0086-3}.
\newblock \urlprefix\url{https://doi.org/10.1007/s12065-012-0086-3}

\bibitem{eustaquio_fuzzy_2018}
Eustáquio, F., Camargo, H., Rezende, S., Nogueira, T.: On {Fuzzy} {Cluster}
  {Validity} {Indexes} for {High} {Dimensional} {Feature} {Space}.
\newblock In: J.~Kacprzyk, E.~Szmidt, S.~Zadrożny, K.T. Atanassov, M.~Krawczak
  (eds.) Advances in {Fuzzy} {Logic} and {Technology} 2017, Advances in
  {Intelligent} {Systems} and {Computing}, pp. 12--23. Springer International
  Publishing (2018)

\bibitem{eustaquio_monotonic_2018}
Eustáquio, F., Nogueira, T.: On {Monotonic} {Tendency} of {Some} {Fuzzy}
  {Cluster} {Validity} {Indices} for {High}-{Dimensional} {Data}.
\newblock In: 2018 7th {Brazilian} {Conference} on {Intelligent} {Systems}
  ({BRACIS}), pp. 558--563 (2018).
\newblock \doi{10.1109/BRACIS.2018.00102}

\bibitem{fabro_using_2014}
Fabro, J.A., Reis, L.P., Lau, N.: Using {Reinforcement} {Learning} {Techniques}
  to {Select} the {Best} {Action} in {Setplays} with {Multiple} {Possibilities}
  in {Robocup} {Soccer} {Simulation} {Teams}.
\newblock In: 2014 {Joint} {Conference} on {Robotics}: {SBR}-{LARS} {Robotics}
  {Symposium} and {Robocontrol}, pp. 85--90. IEEE, Sao Carlos, Sao Paulo,
  Brazil (2014).
\newblock \doi{10.1109/SBR.LARS.Robocontrol.2014.47}.
\newblock \urlprefix\url{http://ieeexplore.ieee.org/document/7024261/}

\bibitem{freelan_towards_2014}
Freelan, D., Wicke, D., Sullivan, K., Luke, S.: Towards {Rapid} {Multi}-robot
  {Learning} from {Demonstration} at the {RoboCup} {Competition}.
\newblock In: {RoboCup} 2014: {Robot} {World} {Cup} {XVIII}, Lecture {Notes} in
  {Computer} {Science}, pp. 369--382. Springer, Cham (2014).
\newblock \doi{10.1007/978-3-319-18615-3_30}.
\newblock
  \urlprefix\url{https://link.springer.com/chapter/10.1007/978-3-319-18615-3_30}

\bibitem{hieu_cell-mst-based_2015}
Hieu, D.V., Meesad, P.: A cell-{MST}-based method for big dataset clustering on
  limited memory computers.
\newblock In: 2015 7th {International} {Conference} on {Information}
  {Technology} and {Electrical} {Engineering} ({ICITEE}), pp. 632--637 (2015).
\newblock \doi{10.1109/ICITEED.2015.7409023}

\bibitem{hoppner_fuzzy_1999}
Höppner, F., Klawonn, F., Kruse, R., Runkler, T.: Fuzzy {Cluster} {Analysis}:
  {Methods} for {Classification}, {Data} {Analysis} and {Image} {Recognition}.
\newblock John Wiley \& Sons (1999).
\newblock Google-Books-ID: ZWaREPjUVeMC

\bibitem{liemhetcharat_allocating_2017}
Liemhetcharat, S., Veloso, M.: Allocating training instances to learning agents
  for team formation.
\newblock Autonomous Agents and Multi-Agent Systems \textbf{31}(4), 905--940
  (2017).
\newblock \doi{10.1007/s10458-016-9355-3}.
\newblock \urlprefix\url{https://doi.org/10.1007/s10458-016-9355-3}

\bibitem{micalizio_explaining_2016}
Micalizio, R., Torta, G.: Explaining interdependent action delays in multiagent
  plans execution.
\newblock Autonomous Agents and Multi-Agent Systems \textbf{30}(4), 601--639
  (2016).
\newblock \doi{10.1007/s10458-015-9298-0}.
\newblock \urlprefix\url{https://doi.org/10.1007/s10458-015-9298-0}

\bibitem{moradi_learning_2016}
Moradi, M., Ardestani, M.A., Moradi, M.: Learning decision making for {Soccer}
  {Robots}: {A} crowdsourcing-based approach.
\newblock In: 2016 {Artificial} {Intelligence} and {Robotics} ({IRANOPEN}), pp.
  25--29 (2016).
\newblock \doi{10.1109/RIOS.2016.7529514}

\bibitem{hutchison_collaborative_2015}
Mota, L., Fabro, J.A., Reis, L.P., Lau, N.: Collaborative {Behavior} in
  {Soccer}: {The} {Setplay} {Free} {Software} {Framework}.
\newblock In: D.~Hutchison, T.~Kanade, J.~Kittler, J.M. Kleinberg, F.~Mattern,
  J.C. Mitchell, M.~Naor, O.~Nierstrasz, C.~Pandu~Rangan, B.~Steffen, M.~Sudan,
  D.~Terzopoulos, D.~Tygar, M.Y. Vardi, G.~Weikum, F.~Bao, P.~Samarati, J.~Zhou
  (eds.) Applied {Cryptography} and {Network} {Security}, vol. 7341, pp.
  709--716. Springer Berlin Heidelberg, Berlin, Heidelberg (2015).
\newblock \doi{10.1007/978-3-319-18615-3_58}.
\newblock \urlprefix\url{http://link.springer.com/10.1007/978-3-319-18615-3_58}

\bibitem{mota_co-ordination_2010}
Mota, L., Lau, N., Reis, L.P.: Co-ordination in {RoboCup}'s 2d simulation
  league: {Setplays} as flexible, multi-robot plans.
\newblock In: 2010 {IEEE} {Conference} on {Robotics}, {Automation} and
  {Mechatronics}, pp. 362--367 (2010).
\newblock \doi{10.1109/RAMECH.2010.5513166}

\bibitem{mota_multi-robot_2011}
Mota, L., Reis, L.P., Lau, N.: Multi-robot coordination using {Setplays} in the
  middle-size and simulation leagues.
\newblock Mechatronics \textbf{21}(2), 434--444 (2011).
\newblock \doi{10.1016/j.mechatronics.2010.05.005}.
\newblock
  \urlprefix\url{https://linkinghub.elsevier.com/retrieve/pii/S0957415810000851}

\bibitem{nayak_fuzzy_2015}
Nayak, J., Naik, B., Behera, H.S.: Fuzzy {C}-{Means} ({FCM}) {Clustering}
  {Algorithm}: {A} {Decade} {Review} from 2000 to 2014.
\newblock In: L.C. Jain, H.S. Behera, J.K. Mandal, D.P. Mohapatra (eds.)
  Computational {Intelligence} in {Data} {Mining} - {Volume} 2, pp. 133--149.
  Springer India (2015)

\bibitem{panella_interactive_2017}
Panella, A., Gmytrasiewicz, P.: Interactive {POMDPs} with finite-state models
  of other agents.
\newblock Autonomous Agents and Multi-Agent Systems \textbf{31}(4), 861--904
  (2017).
\newblock \doi{10.1007/s10458-016-9359-z}.
\newblock \urlprefix\url{https://doi.org/10.1007/s10458-016-9359-z}

\bibitem{ramos_planejador_2017}
Ramos, C.E.d.R.: Planejador {Multiagentes} para {Criação} de {Jogadas}
  {Ensaiadas} em um {Time} de {Futebol} de {Robôs} {Simulados}.
\newblock Bachelor {Thesis}, Universidade do Estado da Bahia (UNEB), Salvador,
  BA, Brazil (2017).
\newblock
  \urlprefix\url{http://www.acso.uneb.br/bahiart/index.php?n=Main.TCC?action=bibentry&bibfile=BahiaRT_TCC.bib&bibref=ReisRamos2017}

\bibitem{reis_situation_2001}
Reis, L.P., Lau, N., Oliveira, E.C.: Situation {Based} {Strategic}
  {Positioning} for {Coordinating} a {Team} of {Homogeneous} {Agents}.
\newblock In: M.~Hannebauer, J.~Wendler, E.~Pagello (eds.) Balancing
  {Reactivity} and {Social} {Deliberation} in {Multi}-{Agent} {Systems},
  Lecture {Notes} in {Computer} {Science}, pp. 175--197. Springer Berlin
  Heidelberg (2001)

\bibitem{reis_playmaker:_2010}
Reis, L.P., Lopes, R., Mota, L., Lau, N.: Playmaker: {Graphical} definition of
  formations and setplays.
\newblock In: 5th {Iberian} {Conference} on {Information} {Systems} and
  {Technologies}, pp. 1--6 (2010)

\bibitem{sardar_evaluation_2017}
Sardar, T.H., Ansari, Z., Khatun, A.: An evaluation of {Hadoop} cluster
  efficiency in document clustering using parallel {K}-means.
\newblock In: 2017 {IEEE} {International} {Conference} on {Circuits} and
  {Systems} ({ICCS}), pp. 17--20 (2017).
\newblock \doi{10.1109/ICCS1.2017.8325954}

\bibitem{shao_clustering_2013}
Shao, W., Shi, X., Yu, P.S.: Clustering on {Multiple} {Incomplete} {Datasets}
  via {Collective} {Kernel} {Learning}.
\newblock In: 2013 {IEEE} 13th {International} {Conference} on {Data} {Mining},
  pp. 1181--1186. IEEE, Dallas, TX, USA (2013).
\newblock \doi{10.1109/ICDM.2013.117}.
\newblock \urlprefix\url{http://ieeexplore.ieee.org/document/6729618/}

\bibitem{shi_adaptive_2018}
Shi, H., Lin, Z., Hwang, K., Yang, S., Chen, J.: An {Adaptive} {Strategy}
  {Selection} {Method} {With} {Reinforcement} {Learning} for {Robotic} {Soccer}
  {Games}.
\newblock IEEE Access \textbf{6}, 8376--8386 (2018).
\newblock \doi{10.1109/ACCESS.2018.2808266}

\bibitem{simoes_towards_2018}
Simões, M.A.C., Nogueira, T.: Towards {Setplays} {Learning} in a {Multiagent}
  {Robotic} {Soccer} {Team}.
\newblock In: 2018 {Latin} {American} {Robotic} {Symposium}, 2018 {Brazilian}
  {Symposium} on {Robotics} ({SBR}) and 2018 {Workshop} on {Robotics} in
  {Education} ({WRE}), pp. 277--282 (2018).
\newblock \doi{10.1109/LARS/SBR/WRE.2018.00058}

\bibitem{stone_task_1999}
Stone, P., Veloso, M.: Task decomposition, dynamic role assignment, and
  low-bandwidth communication for real-time strategic teamwork.
\newblock Artificial Intelligence \textbf{110}(2), 241--273 (1999).
\newblock \doi{10.1016/S0004-3702(99)00025-9}.
\newblock
  \urlprefix\url{https://linkinghub.elsevier.com/retrieve/pii/S0004370299000259}

\bibitem{nguyen_fast_2015}
Van~Hieu, D., Meesad, P.: Fast {K}-{Means} {Clustering} for {Very} {Large}
  {Datasets} {Based} on {MapReduce} {Combined} with a {New} {Cutting} {Method}.
\newblock In: V.H. Nguyen, A.C. Le, V.N. Huynh (eds.) Knowledge and {Systems}
  {Engineering}, vol. 326, pp. 287--298. Springer International Publishing,
  Cham (2015).
\newblock \doi{10.1007/978-3-319-11680-8_23}.
\newblock \urlprefix\url{http://link.springer.com/10.1007/978-3-319-11680-8_23}

\bibitem{wooldridge_introduction_2002}
Wooldridge, M.: An {Introduction} to {Multiagent} {Systems}, 1 edn.
\newblock John Wiley \& Sons, Liverpool (2002)

\bibitem{yang_hybrid_2012}
Yang, Y.S., Li, G., Zhu, Y.S., Zhang, Y.Y.: Hybrid {Genetic} {Clustering} by
  {Using} {FCM} and {Geodesic} {Distance} for {Complex} {Distributed} {Data}.
\newblock Applied Mechanics and Materials; Zurich \textbf{263-266} (2012).
\newblock \doi{http://dx.doi.org/10.4028/www.scientific.net/AMM.263-266.2597}.
\newblock
  \urlprefix\url{https://search.proquest.com/docview/1442793660/abstract/A25A7D33D8ED4F71PQ/1}

\bibitem{yu_multiagent_2015}
Yu, C., Zhang, M., Ren, F., Tan, G.: Multiagent {Learning} of {Coordination} in
  {Loosely} {Coupled} {Multiagent} {Systems}.
\newblock IEEE Transactions on Cybernetics \textbf{45}(12), 2853--2867 (2015).
\newblock \doi{10.1109/TCYB.2014.2387277}

\bibitem{zhang_keeping_2015}
Zhang, C., Sinha, A., Tambe, M.: Keeping {Pace} with {Criminals}: {Designing}
  {Patrol} {Allocation} {Against} {Adaptive} {Opportunistic} {Criminals}.
\newblock In: roceedings of the 2015 {International} {Conference} on
  {Autonomous} {Agents} and {Multiagent} {Systems}, pp. 1351--1359. Istanbul,
  Turkey (2015)

\bibitem{zhang_modeling_2015}
Zhang, C., Tambe, M.: Modeling, {Learning} and {Defending} against
  {Opportunistic} {Criminals} in {Urban} {Areas} ({Doctoral} {Consortium}).
\newblock In: Proceedings of the 2015 {International} {Conference} on
  {Autonomous} {Agents} and {Multiagent} {Systems}, pp. 1971--1972. Istambul,
  Turkey (2015)

\bibitem{zhang_multirobot_2017}
Zhang, S., Jiang, Y., Sharon, G., Stone, P.: Multirobot {Symbolic} {Planning}
  under {Temporal} {Uncertainty}.
\newblock In: Proceedings of the 16th {Conference} on {Autonomous} {Agents} and
  {MultiAgent} {Systems}, pp. 501--510. São Paulo, Brazil (2017)

\bibitem{zhou_combined_2015}
Zhou, J., Purvis, M., Muhammad, Y.: A {Combined} {Modelling} {Approach} for
  {Multi}-{Agent} {Collaborative} {Planning} in {Global} {Supply} {Chains}.
\newblock In: 2015 8th {International} {Symposium} on {Computational}
  {Intelligence} and {Design} ({ISCID}), vol.~1, pp. 592--597 (2015).
\newblock \doi{10.1109/ISCID.2015.13}

\bibitem{ziani_feature_2014}
Ziani, D.: Feature selection on probabilistic symbolic objects.
\newblock Frontiers of Computer Science \textbf{8}(6), 933--947 (2014).
\newblock \doi{10.1007/s11704-014-3359-4}.
\newblock \urlprefix\url{https://doi.org/10.1007/s11704-014-3359-4}

\end{thebibliography}
